\documentclass[twoside]{article}
\usepackage[accepted]{aistats2021}

\usepackage{amsmath}
\usepackage{amssymb}
\usepackage{commath}

\usepackage{booktabs}
\usepackage{multirow}
\usepackage{pgfplots}

\usepackage[round]{natbib}

\usepackage{hyperref}
\usepackage[all]{hypcap}
\usepackage{xcolor}
\hypersetup{
    colorlinks,
    linkcolor={red!50!black},
    citecolor={red!50!black},
    urlcolor={red!50!black}
}

\pgfplotsset{compat=1.14}

\newcommand{\lam}{\lambda}
\newcommand{\reals}{\mathbb{R}}
\newcommand{\normal}{\mathcal{N}}

\newcommand{\expect}{\mathbb{E}}

\newcommand{\half}{\tfrac{1}{2}}

\DeclareMathOperator{\KL}{KL}
\DeclareMathOperator{\JS}{JS}

\newcommand{\divergence}[2]{(#1\,\|\,#2)}
\newcommand{\KLD}[2]{\KL\divergence{#1}{#2}}

\newcommand{\sref}[1]{\S\ref{#1}}
\newcommand{\tabref}[1]{Table~\ref{#1}}
\newcommand{\Eqref}[1]{(\ref{#1})}

\begin{document}

\twocolumn[

\aistatstitle{Non-saturating GAN training as divergence minimization}

\aistatsauthor{Matt Shannon
  \And Ben Poole
  \And Soroosh Mariooryad
  \And Tom Bagby
}
\aistatsauthor{Eric Battenberg
  \And David Kao
  \And Daisy Stanton
  \And RJ Skerry-Ryan
}

\aistatsaddress{Google, Mountain View, California, USA}

]  

\begin{abstract}
Non-saturating generative adversarial network (GAN) training is widely used and has continued to
obtain groundbreaking results.
However so far this approach has lacked strong theoretical justification, in contrast to alternatives
such as f-GANs and Wasserstein GANs which are motivated in terms of approximate divergence minimization.
In this paper we show that non-saturating GAN training does in fact approximately minimize a particular
f-divergence.
We develop general theoretical tools to compare and classify f-divergences and use these to show
that the new f-divergence is qualitatively similar to reverse KL.
These results help to explain the high sample quality but poor diversity often observed
empirically when using this scheme.
\end{abstract}

\section{Introduction}
\label{sec:intro}
Generative adversarial networks (GANs) \citep{goodfellow_generative_2014} have enjoyed
remarkable progress in recent years, producing images of striking fidelity, resolution
and coherence \citep{karras2018progressive,miyato2018spectral,brock2018large,karras2019style}.
A GAN consists of two components:
a \emph{generator} produces a synthetic image (for example) and a \emph{critic} or
\emph{discriminator} aims to distinguish these synthetic images from natural ones.
The generator and critic are trained adversarially to try to outwit each other, thus
improving the generator.
There has been recent progress in both theoretical and practical aspects of understanding and
performing GAN training
\citep{nowozin_f-gan:_2016,arjovsky_towards_2017,arjovsky_wasserstein_2017,mescheder2018training,%
gulrajani2017improved,sonderby2017amortised,%
miyato2018spectral,karras2018progressive,brock2018large,karras2019style}.

A rich vein of developments has come from viewing GAN training as \emph{divergence minimization}.
The discrepancy between the data distribution and the distribution of generator output is measured
by a probabilistic divergence.
The critic has an ancillary role, allowing this divergence to be tractably estimated and minimized.
Conventional GAN training approximately minimizes the Jensen-Shannon divergence
\citep{goodfellow_generative_2014},
f-GANs \citep{nowozin_f-gan:_2016} approximately minimize f-divergences such as reverse KL,
and Wasserstein GANs \citep{arjovsky_wasserstein_2017} approximately minimize the Wasserstein-1 metric.

Despite these theoretically well-founded developments, one of the most widely used GAN
training methods is a heuristic \emph{non-saturating} scheme \citep{goodfellow_generative_2014}.
This modifies the original Jensen-Shannon-based generator gradient to provide a stronger
learning signal early in training.
This approach has continued to obtain groundbreaking results \citep{karras2019style}
and outperformed other GAN variants on some tasks in a recent systematic comparison
\citep{lucic2018gans}.

The main result of this paper is that the non-saturating scheme approximately minimizes the
f-divergence $\KLD{\half p + \half q}{p}$, which we call the \emph{softened reverse KL}
(\sref{sec:equiv-ns}).
This puts the non-saturating scheme on a similar footing to Wasserstein GANs as a principled
approach with strong empirical results.

Our second major contribution is a set of general tools to compare f-divergences.
We show how to write f-divergences in a way that allows easy visual comparison
of their qualitative properties (\sref{sec:pushforward-symm}), and develop a formulation
of \emph{tail weight} which generalizes the notions of \emph{mode-seeking} and \emph{covering}
behavior (\sref{sec:tail-weights}).

We use these general tools to show that softened reverse KL is mode-seeking and is
qualitatively similar to reverse KL (\sref{sec:effect-ns}),
helping to explain the high sample quality but poor sample diversity of most GAN models trained
with the non-saturating scheme.

Our final major contribution is to clarify the substantial confusion and conflicting claims
in previous literature on theoretical aspects of non-saturating training (\sref{sec:previous-work}).

\section{f-GAN training}
\label{sec:f-gan-training}
In this section we review f-GAN training \citep{nowozin_f-gan:_2016}.
We use a slightly modified formulation, presented in detail in a technical report
\citep{shannon2020properties},
which ensures that the optimal critic is the same for all f-divergences and avoids
having to worry about various small details such as the domains of Fenchel conjugates
\citep{nowozin_f-gan:_2016}.
Throughout the paper we use the convention that $p$ is the ``true'' distribution and
the \emph{generator} distribution $q$ is a model intended to approximate $p$.

\subsection{f-GAN formulation}
\label{sec:f-gan-formulation}
Given a strictly convex twice continuously differentiable function
$f : \reals_{> 0} \to \reals$ with $f(1) = f'(1) = 0$, the \emph{f-divergence}
\citep{csiszar1967information,ali1966general}
between probability distributions with densities $p$ and $q$ over $\reals^K$ is
defined as
\begin{equation}
  \label{eq:f-def}
  D_f(p, q) = \expect_{x \sim q}\left[ f\left(\frac{p(x)}{q(x)}\right) \right]
\end{equation}
For simplicity, we assume the probability distributions $p$ and $q$ are suitably nice,
e.g.\ absolutely continuous with respect to the Lebesgue measure on $\reals^K$,
$p(x), q(x) > 0$ for $x \in \reals^K$, and $p$ and $q$ continuously differentiable.%
\footnote{%
  The constraint $q(x) > 0$ is often violated in practice due to the GAN generator
  taking a low-dimensional random input to produce a high-dimensional output.
  We advocate injecting noise at all levels of the generator network including the output
  \citep{shannon2020properties}.
  This can have advantages in practice \citep{karras2019style}.
}
We refer to $f$ as the \emph{defining function} of the divergence $D_f$.
The constraint $f'(1) = 0$ removes an irrelevant degree of freedom and is not a restriction
\citep{shannon2020properties}.
Since $f(1) = f'(1) = 0$, $f''$ completely determines $f$.
Working in terms of $f''$ is convenient since $f''$ has a simpler algebraic form than
$f$ for many common f-divergences.

An f-divergence has a simple variational lower bound based on bounding $f$ by its tangent lines
\citep{nowozin_f-gan:_2016,shannon2020properties}.
For any continuously differentiable function $d : \reals^K \to \reals$, we have
\begin{equation*}
  D_f(p, q) \geq E_f(p, q, d)
\end{equation*}
with equality iff $d = d^*$, where
\begin{align}
  \label{eq:f-bound-def}
  E_f(p, q, d)
    &= \expect_{x \sim p}\left[ a_f(d(x)) \right] - \expect_{x \sim q}\left[ b_f(d(x)) \right]
  \\
  d^*(x) &= \log p(x) - \log q(x)
\end{align}
and $a_f, b_f : \reals \to \reals$ are defined by
\begin{align}
  a_f(\log u) &= f'(u)
  \\
  b_f(\log u) &= u f'(u) - f(u)
\end{align}
The \emph{critic} or \emph{discriminator} $d$ approximates the log density ratio between $p$
and $q$.

Typically the generator distribution and critic are both parameterized.
The critic is parameterized directly as a neural net $d_\nu$ with parameters $\nu$.
The generator distribution is specified implicitly: noise $z$ sampled from a fixed
distribution is passed through a neural net $g_\lam$ with parameters $\lam$ and the
output $g_\lam(z)$ is taken as a sample from $q_\lam$.
The value $E_f(p, q_\lam, d_\nu)$ and its gradient with respect to the critic parameters
$\nu$ can be approximated using samples from $p$ and $q_\lam$.
To compute the gradient with respect to the generator parameters $\lam$, it is helpful to
``reparameterize'' the expectation over $q$ in \Eqref{eq:f-bound-def} to obtain
\citep{kingma2014autoencoding}
\begin{equation}
  \label{eq:reparam-generator-loss}
  E_f(p, q_\lam, d) = -\expect_z\left[ b_f(d(g_\lam(z))) \right] + \text{const}
\end{equation}

Training of the (critic, generator) system is inherently adversarial, since obtaining an accurate
estimate of $D_f(p, q_\lam)$ requires maximizing the lower bound $E_f(p, q_\lam, d_\nu)$
with respect to $\nu$, but overall we want to minimize $D_f(p, q_\lam)$ with respect to $\lam$.
Typically both $\lam$ and $\nu$ are updated using a gradient-based optimizer such as
SGD, ADAM or RMSProp, with a single generator update performed after or simultaneously with
one or more critic updates.

Convergence of GAN training algorithms is still a complicated topic despite much attention
\citep[for example]{nagarajan2017gradient,gulrajani2017improved,%
kodali2017convergence,%
mescheder2017numerics,mescheder2018training,%
fedus_many_2018,%
balduzzi2018mechanics,peng2019training}.
The definition of $E_f(p, q_\lam, d)$ ensures that, in the case where the critic $d$ is
unrestricted, the only fixed points of training dynamics are at $(\lam, d) = (\lam^*, 0)$
for $\lam^*$ a stationary point of $\lam \mapsto D_f(p, q_\lam)$.
These fixed points are often Nash equilibria and locally stable
\citep{nagarajan2017gradient,mescheder2018training}.
The gradient matching property described in \sref{sec:f-gan-properties} shows that
performing very many critic updates from scratch for each generator update, but not so
many as to overfit, essentially performs gradient descent on $D_f$, and so will find
a local minimum in $D_f$.
However in practice the critic will always be somewhat suboptimal and the generator
updates will not follow the gradient of $D_f$ exactly (see \sref{sec:experiments} for a
simple example).
The restricted parametric form of the critic may also lead to suboptimal fixed points.
Because of the issues with parameterization and optimization of the critic, we refer
to f-GAN training as \emph{approximate} divergence minimization.

The three main f-divergences we consider are:
\begin{itemize}
\item 
  The Kullback-Leibler (KL) divergence $\KLD{p}{q}$.
  This has $f''(u) = u^{-1}$, $a_f'(d) = 1$ and $b_f'(d) = \exp d$.
\item
  The reverse KL divergence $\KLD{q}{p}$.
  This has $f''(u) = u^{-2}$, $a_f'(d) = \exp(-d)$ and $b_f'(d) = 1$.
\item
  The canonicalized (see \sref{sec:f-gan-properties}) Jensen-Shannon divergence
  $4 \JS(p, q) = 2 \KLD{p}{m} + 2 \KLD{q}{m}$ where $m = \half p + \half q$.
  This has $f''(u) = \frac{2}{u (u + 1)}$,
  $a_f'(d) = 2 \sigma(-d)$ and $b_f'(d) = 2 \sigma(d)$,
  where $\sigma$ is the logistic sigmoid function $\sigma(d) = 1 / (1 + \exp(-d))$.
\end{itemize}

\subsection{Properties of f-GAN training}
\label{sec:f-gan-properties}
In this section we describe three properties of f-GAN training which are relevant
to our discussion of the non-saturating loss in \sref{sec:ns} and \sref{sec:discussion}.

Different f-divergences may behave very differently when $p$ and $q$ are far apart but
are essentially identical when $p$ is close to $q$:
if $p \approx q$ then $D_f(p, q) \approx f''(1) \KLD{p}{q}$ to second order
\citep{sason2016f,shannon2020properties}.
The factor $f''(1)$ represents an overall scale factor which is often not important, and
so we typically scale f-divergences to have $f''(1) = 1$, which we refer to as
\emph{canonical} form.
Thus all canonical f-divergences agree when $p$ is close to $q$.

The variational lower bound $E_f$ satisfies a convenient gradient matching property
not mentioned by \citet{nowozin_f-gan:_2016}.
We have already seen that the values of $E_f$ and $D_f$ match when the critic is optimal.
The same is true of the generator gradient:
\begin{equation}
  \label{eq:grad-match}
  \dpd{}{\lam} D_f(p, q_\lam) =
    \dpd{}{\lam} E_f(p, q_\lam, d)\sVert[4]_{d = d^*_\lam}
\end{equation}
This can be derived from $E_f$ being a tight lower bound or directly from the
definition of $E_f$ \citep{shannon2020properties}.
\citet[Section 4.2]{goodfellow_generative_2014} described this property for the
specific case of Jensen-Shannon.
Wasserstein GANs satisfy a similar property \citep{arjovsky_wasserstein_2017}.
As far as we are aware, this result is novel for general f-GANs.
The gradient matching property shows that the best estimate of the generator gradient for a
given $\lam$ is obtained by having as well-trained a critic as possible.
Any weak generator gradient when using a strong critic is due to the original f-divergence
having a weak gradient (e.g.\ Jensen-Shannon).

There is a simple generalization of the above training procedure, which is to base the
generator gradient on $E_f$ but the critic gradient on $E_h$ for a possibly different
defining function $h$ \citep[Section 2.3]{poole2016improved}.
We refer to this as a \emph{hybrid $(f, h)$} f-GAN training scheme.
The optimal critic for $h$ is the same as the optimal critic for $f$.
Therefore, in the case where the critic $d$ is unrestricted, the only fixed points of training
dynamics are again at $(\lam, d) = (\lam^*, 0)$ for $\lam^*$ a stationary point of
$\lam \mapsto D_f(p, q_\lam)$.
The same argument about behavior in the limit of a large number of critic updates per
generator update also applies.
Thus a hybrid $(f, h)$ scheme may be interpreted as minimizing $D_f$, the divergence used
for updating the \emph{generator}.

\section{Non-saturating GAN training}
\label{sec:ns}
In practice the original generator gradient based on the Jensen-Shannon divergence performs
poorly, and typically an alternative \emph{non-saturating} generator gradient is used instead
\citep{goodfellow_generative_2014}.
In this section we briefly review the issue with the original generator gradient, describe the
non-saturating fix, and establish our main result relating non-saturating training to hybrid
f-GAN training.

\subsection{Non-saturating training procedure}
Early on in training, the generator and data distribution are typically not well matched,
with samples from $p$ being very unlikely under $q$ and vice versa.
This means most of the probability mass of $p$ and $q$ is in regions where $d$ has large
magnitude (corresponding to the positive and negative tails in \figref{fig:symm} and
\Eqref{eq:symm-f-def} below).
In this regime Jensen-Shannon essentially saturates at its maximum value, and it is not
too surprising that this might lead to optimization issues.
Similar concerns do not apply to other f-divergences such as KL or reverse KL, but an
alternative ``non-saturating'' generator gradient has still been suggested for f-GANs
\citep{nowozin_f-gan:_2016}.
For both GANs and f-GANs the specific change is to replace $b_f$ by $a_f$ when computing the
generator gradient using \Eqref{eq:reparam-generator-loss}, that is to use
\begin{equation}
  \label{eq:reparam-generator-ns-loss}
  E_f(p, q_\lam, d) = -\expect_z\left[ a_f(d(g_\lam(z))) \right] + \text{const}
\end{equation}
In the case of canonicalized Jensen-Shannon (i.e.\ conventional non-saturating GAN training)
this means replacing
$b_f(d) = -2 \log \sigma(-d) - 2 \log 2$ with $a_f(d) = 2 \log \sigma(d) + 2 \log 2$.
We are not aware of a particular motivation for this procedure in the case of f-GANs other
than that it yields the traditional non-saturating GAN scheme in the case of Jensen-Shannon.

\subsection{Equivalence to hybrid f-GAN training}
\label{sec:equiv-ns}
We now establish our main result:
for any f-divergence $D_h$, ``non-saturating'' training based on $h$ is precisely equivalent
to a hybrid $(f, h)$ scheme for some defining function $f$.
This shows that the non-saturating scheme is not simply a trick to get a useful gradient,
but entirely changes the divergence minimized by training.

Given $h$, we seek a defining function $f$ such that $b_f' = a_h'$, since then the original
generator gradient based on $f$ obtained by differentiating \Eqref{eq:reparam-generator-loss}
will be equal to the ``non-saturating'' generator gradient based on $h$ obtained by
differentiating \Eqref{eq:reparam-generator-ns-loss}.
Since $a_h'(\log u) = u h''(u)$ and $b_f'(\log u) = u^2 f''(u)$, there is only one
possibility:
\begin{equation}
  \label{eq:ns-equiv}
  f''(u) = u^{-1} h''(u)
\end{equation}
This defines a valid f-divergence since $f''(u) > 0$ and satifies $b_f' = a_h'$ as desired.
Since the critic gradient is still based on $h$, the overall scheme is a
hybrid $(f, h)$ one, and so approximately minimizes $D_f$.

We now explicitly compute the corresponding $f$ for some common choices of $h$:
\begin{itemize}
\item
  For the KL divergence, $h''(u) = u^{-1}$, so $f''(u) = u^{-2}$.
  We already saw in \sref{sec:f-gan-formulation} that this is the reverse KL divergence.
  Thus ``non-saturating'' training based on the KL divergence is a hybrid
  (reverse KL, KL) scheme, and so in fact approximately minimizes the reverse KL.
  This equivalence also follows directly from the equality of KL's $a_h$ to reverse KL's
  $b_f$.
\item
  For the reverse KL divergence, $h''(u) = u^{-2}$, so $f''(u) = u^{-3}$.
  Integrating twice to obtain $f$, this corresponds to the canonicalized Neymann $\chi^2$
  divergence $D_f(p, q) = \half \int (q(x) - p(x))^2 / p(x) \dif x$.
\item
  For the canonicalized Jensen-Shannon divergence, $h''(u) = \frac{2}{u (u + 1)}$, so
  $f''(u) = \frac{2}{u^2 (u + 1)}$.
  Integrating twice to obtain $f$, this corresponds to
  $D_f(p, q) = 4 \KLD{\half p + \half q}{p}$.
\end{itemize}
The divergence $4 \KLD{\half p + \half q}{p}$
does not have an existing name as far as we are aware.
In this paper we have termed it the \emph{softened reverse KL (SRKL)}.
The naming is explained in \sref{sec:softening}.
It has $a_f'(d) = 2 \exp(-d) - 2 \sigma(-d)$ and $b_f'(d) = 2 \sigma(-d)$.
Thus the non-saturating training scheme described by \citet{goodfellow_generative_2014}
is a hybrid (SRKL, JS) scheme, and so approximately minimizes the softened reverse KL.

In \sref{sec:experiments} we validate our mathematical conclusions by training a
GAN for which the evolution of the generator parameters can be directly observed.
As expected, conventional saturating GAN training converges to a minimum of the
Jensen-Shannon divergence and non-saturating GAN training converges to a minimum
of the softened reverse KL.

\section{Tools to compare f-divergences}
\label{sec:tools}
Having established that non-saturating GAN training approximately minimizes the
softened reverse KL divergence, we focus on understanding the qualitative
properties of this divergence.
In this section we develop analytic tools applicable to any f-divergence.
We apply these tools to the softened reverse KL in \sref{sec:effect-ns}.

\subsection{Left and right mismatches}
\label{sec:mismatches}
We start by developing a more symmetric representation of f-divergences.
The definition \Eqref{eq:f-def} appears to be quite asymmetric in how it treats $p$ and $q$,
but it obeys a particular symmetry \citep{reid2011information}.
Define the \emph{Csziz{\'a}r dual} $f_\text{R}$ by $f_\text{R}(u) = u f(u^{-1})$.
Then $f_\text{R}'(u) = f(u^{-1}) - u^{-1} f'(u^{-1})$ and so
$f_\text{R}''(u) = u^{-3} f''(u^{-1})$.
It is easy to verify that $D_{f_\text{R}}(p, q) = D_f(q, p)$.
With $A = \{x : q(x) > p(x)\}$ and $B = \{x : q(x) < p(x)\}$, we have
\begin{equation}
  \label{eq:f-left-right-decomp}
  D_f(p, q)
    = \int_A q(x) f\left(\frac{p(x)}{q(x)}\right) \dif x
      + \int_B p(x) f_\text{R}\left(\frac{q(x)}{p(x)}\right) \dif x
\end{equation}
This is more explicitly symmetric than \Eqref{eq:f-def} in the role of $p$ and $q$.
We refer to $A$ as the set of \emph{left mismatches} and $B$ as the set of
\emph{right mismatches}.
At each point in $A$, the model assigns higher density than the data, and the
penalty paid for this mismatch in terms of the overall divergence $D_f$ is governed by the
behavior of $f(u)$ for $0 < u < 1$ (the ``left'' of the graph of $f$).
Similarly the penalty paid for right mismatches, where the model assigns lower density than
the data, is governed by $f(u)$ for $u > 1$.
Note from \Eqref{eq:f-left-right-decomp} that a left mismatch can only be heavily penalized
if the point is plausible under $q$, i.e.\ $q(x)$ is not tiny.
Similarly a right mismatch can only be heavily penalized for points plausible under $p$.

\subsection{Pushforwards and symmetry-preserving divergence plots}
\label{sec:pushforward-symm}
While the f-divergence framework unifies many divergences, just plotting the defining
function $f$ is often not informative.
The symmetric relationship between divergences such as KL and reverse KL is obfuscated, and
$f$ may grow quickly even when the divergence is well-behaved.
In this section we develop a straightforward and intuitive way to visually compare f-divergences.

Firstly note that for $x \sim q$, $d^*(x) = \log p(x) - \log q(x)$ is a random variable
with some density $\tilde{q}(d)$ for $d \in \reals$.
Formally $\tilde{q}$ is the \emph{pushforward measure} of $q$ through the function $d^*$.
We can rewrite \Eqref{eq:f-def} as
\begin{equation}
  D_f(p, q) = \expect_{d \sim \tilde{q}}\left[ f(\exp d) \right]
\end{equation}
As above this can be written more symmetrically.
Let
\begin{equation}
    \label{eq:s-def}
    s_f(d) =
    \begin{cases}
      f(\exp d), \quad &d < 0
      \\
      f_\text{R}\left(\exp (-d)\right) = \frac{f(\exp d)}{\exp d}, \quad &d > 0
    \end{cases}
\end{equation}
By considering expectation of an arbitrary function of $d$ expressed in $x$-space and $d$-space, we
can show that
\begin{equation}
  \label{eq:pushforward-relationship}
  \tilde{q}(d) = \tilde{p}(d) \exp(-d)
\end{equation}
where $\tilde{p}$ is the pushforward of $p$ through $d^*$.
Thus, using \Eqref{eq:f-left-right-decomp} and \Eqref{eq:pushforward-relationship}, we can write the
f-divergence as
\begin{align}
  D_f(p, q)
    &= \int_{-\infty}^0 \tilde{q}(d) s_f(d) \dif d
      + \int_0^\infty \tilde{p}(d) s_f(d) \dif d
  \\
  \label{eq:symm-f-def}
    &= \int_{-\infty}^\infty \max\left\{ \tilde{p}(d), \tilde{q}(d) \right\} s_f(d) \dif d
\end{align}
where the last equality follows from \Eqref{eq:pushforward-relationship} since
$\tilde{q}(d)$ is larger than $\tilde{p}(d)$ for $d < 0$ and vice versa for $d > 0$.

\begin{figure}[t]
  \centering
  \includegraphics[trim=5mm 2mm 13mm 10mm,clip,scale=0.58]{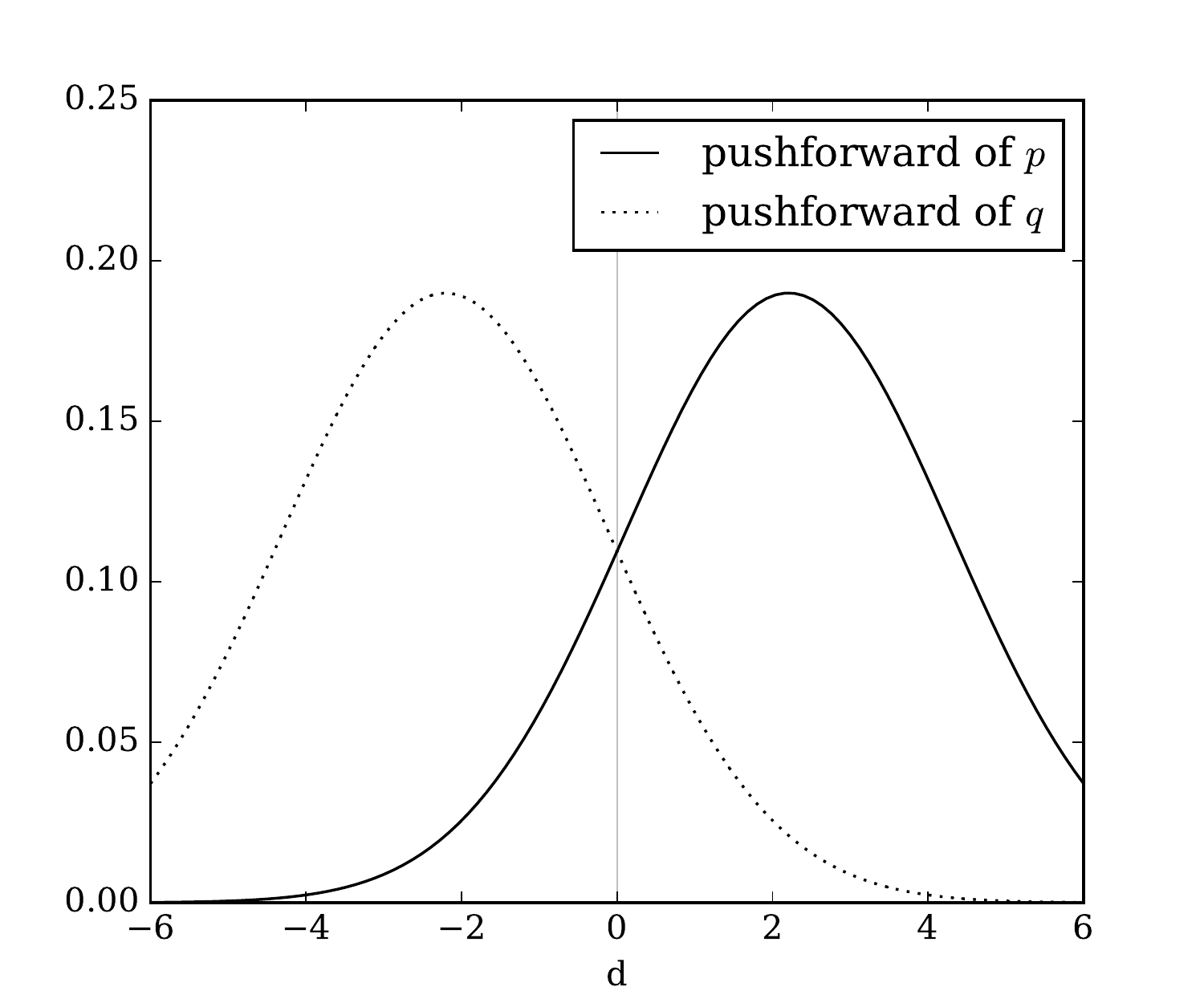}
  \caption{%
    Plots of the pushforward densities $\tilde{p}(d)$ and $\tilde{q}(d)$ for the case where
    $p$ and $q$ are multidimensional Gaussians with common covariance.
    The f-divergence $D_f(p, q)$ may be obtained by integrating these pushforwards against $s_f$
    (e.g.\ \figref{fig:symm}) using \Eqref{eq:symm-f-def}.
  }%
  \label{fig:pushforward-gaussian}
\end{figure}
Examples of pushforwards for the simple case where $p$ and $q$ are multidimensional Gaussians with common
covariance are shown in \figref{fig:pushforward-gaussian}.
In this case the pushforwards $\tilde{q}$ and $\tilde{p}$ are themselves one-dimensional Gaussians
(since $d^*$ is linear), with densities $\normal(-\half \sigma^2, \sigma^2)$ and $\normal(\half \sigma^2, \sigma^2)$
respectively, for some $\sigma$ (this follows from \Eqref{eq:pushforward-relationship}).
\begin{figure}[t]
  \centering
  \includegraphics[trim=15mm 2mm 13mm 10mm,clip,scale=0.58]{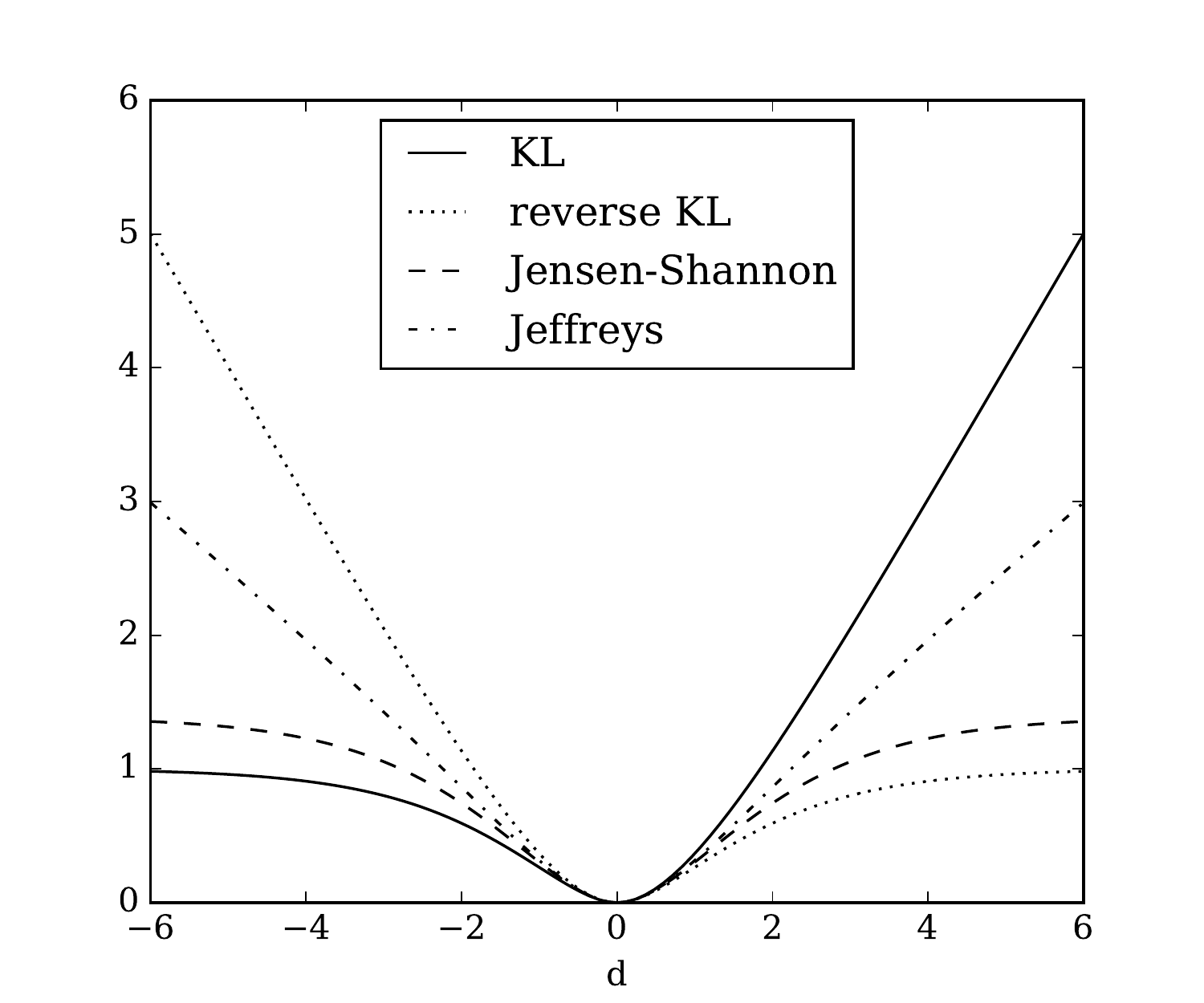}
  \caption{%
    Plots of $s_f(d)$ for various f-divergences.
    The f-divergence $D_f(p, q)$ may be obtained by integrating $s_f$ against their pushforwards
    (e.g.\ \figref{fig:pushforward-gaussian}) using \Eqref{eq:symm-f-def}.
    The symmetry between KL and reverse KL is evident.
  }%
  \label{fig:symm}
\end{figure}
Examples of $s_f$ for various f-divergences are shown in \figref{fig:symm}.
We refer to $s_f$ as a \emph{symmetry-preserving} representation of $f$.
Note that $s_f$ is twice continuously differentiable at zero (using $f'(1) = 0$).

An f-divergence $D_f(p, q)$ involves an interaction between the distributions $p$, $q$ and the
defining function $f$, and \Eqref{eq:symm-f-def} nicely decomposes this interaction in terms of
something that only depends on $p$ and $q$ (the pushforwards) and something that only depends
on $f$ (the function  $s_f$), connected via a one-dimensional integral.
By plotting the pushforwards, we can get a feel for what types of mismatch between $p$ and $q$ are
present in multidimensional $x$-space, and understand at a glance how badly these mismatches would
be penalized for a given f-divergence.
By plotting $s_f$ and imagining integrating against various pushforwards, we can see the properties
of different f-divergences in a very direct way.
For example, \figref{fig:symm} directly expresses several facts about divergences.
It shows that left mismatches (regions of space where $q(x) > p(x)$, corresponding to $d < 0$)
are penalized by reverse KL much more severely than right mismatches (regions of space where
$q(x) > p(x)$, corresponding to $d > 0$).
The symmetry between KL and reverse KL is evident.
We see that Jensen-Shannon and the Jeffreys divergence (the arithmetic mean of KL and reverse KL)
are both symmetric in how they penalize left and right mismatches, but differ greatly in how much
they penalize small versus large mismatches.

\subsection{Classification of f-divergence tails}
\label{sec:tail-weights}
In this section we introduce a classification scheme for f-divergences in terms of their
behavior for large left and right mismatches.
These \emph{tail weights} determine many aspects of an f-divergence's qualitative behavior.

First we define the notion of tail weight and examine some of its consequences.
We write $g(u) \sim h(u)$ as $u \to a$ to mean $g(u) / h(u) \to 1$
as $u \to a$.
If $f''(u) \sim C u^{-L}$ as $u \to 0$ and $f''(u) \sim D u^{R - 3}$ as
$u \to \infty$ where $C, D > 0$ then we say that $D_f$ has
$(C u^{-L}, D u^{R - 3})$ tails and $(L, R)$ tail weights.
Note that, since $f_\text{R}''(u) = u^{-3} f''(u^{-1})$ (see \sref{sec:mismatches}),
$f$ having a $u^{R - 3}$ right tail is equivalent to $f_\text{R}$ having a $u^{-R}$
left tail.
Thus tail weights interact simply with symmetry:
if $D_f(p, q)$ has $(L, R)$ tail weights then $D_{f_\text{R}}(p, q) = D_f(q, p)$ has
$(R, L)$ tail weights.
Intuitively, the left tail weight $L$ determines how strongly large left mismatches are
penalized compared to small mismatches (which are penalized the same amount by every
canonical f-divergence), whereas the right tail weight $R$ determines how strongly large
right mismatches are penalized compared to small mismatches.

\begin{table}[t]
  \centering
  \begin{tabular}{l c c}
    \toprule
    \multirow{2}{*}{divergence} & (left, right) & \multirow{2}{*}{bounded?} \\
    & tail weights & \\
    \midrule
    KL & $(1, 2)$ & no \\
    RKL & $(2, 1)$ & no \\
    Jensen-Shannon & $(1, 1)$ & yes \\
    Jeffreys & $(2, 2)$ & no \\
    Neymann $\chi^2$ & $(3, 0)$ & no \\
    softened RKL & $(2, 0)$ & no \\
    IGOG & $(2, 0)$ & no \\
    \bottomrule
  \end{tabular}
  \caption{%
    Tail weights and boundedness for the f-divergences considered in this paper.
    A divergence is bounded if and only if the left and right tail weights are both
    less than $2$.
  }%
  \label{tab:tail-weights}
\end{table}
Some f-divergences such as Jensen-Shannon are bounded, meaning there is an $M \in \reals$
such that $D_f(p, q) \leq M$ for all densities $p$ and $q$, while others such as KL are
unbounded.
We show in \sref{sec:tail-weights-more} that tail weights determine boundedness:
a divergence is bounded iff $L, R < 2$.
Furthermore $D_f$ is bounded iff $s_f$ is bounded, so we can visually see boundedness on
symmetry-preserving divergence plots such as \figref{fig:symm}.
The tail weights and boundedness properties of various f-divergences are summarized in
\tabref{tab:tail-weights}.

Tail weights provide an extension of the typical classification of divergences as
\emph{mode-seeking} or \emph{covering} \citep[Section 10.1.2]{bishop2006pattern}.
Models trained with reverse KL tend to have distributions which are more ``compact'' or
``localized'' than the true distribution, sometimes only successfully modeling certain modes
(density peaks) of a multi-modal true distribution.
Models trained with KL tend to have distributions which are less compact than the true
distribution, ``covering'' the true distribution entirely even if it means putting density
in regions which are very unlikely under the true distribution
\citep[Figure 10.3]{bishop2006pattern}.
However there are important qualitative aspects of divergence behavior that are not
captured by these labels.
For example, Jensen-Shannon and the Jeffreys divergence are both neither mode-seeking nor
covering, but have very different behavior from each other.
Tail weights capture these distinctions in a straightforward but precise way.

\section{Discussion}
\label{sec:discussion}
In this section we
apply the tools developed in \sref{sec:tools} to
understand the qualitative behavior of the softened reverse KL divergence,
discuss the relationship between our results and previous attempts to
analyze non-saturating GAN training theoretically, and
outline ways the perspectives and tools developed in the paper may be useful in practice.

\subsection{Qualitative properties of softened RKL}
In this section we apply the tools developed in \sref{sec:tools} to analyze the properties
of the softened reverse KL divergence minimized by non-saturating GAN training.

\label{sec:effect-ns}
\begin{figure}[t]
  \centering
  \includegraphics[trim=15mm 2mm 13mm 10mm,clip,scale=0.58]{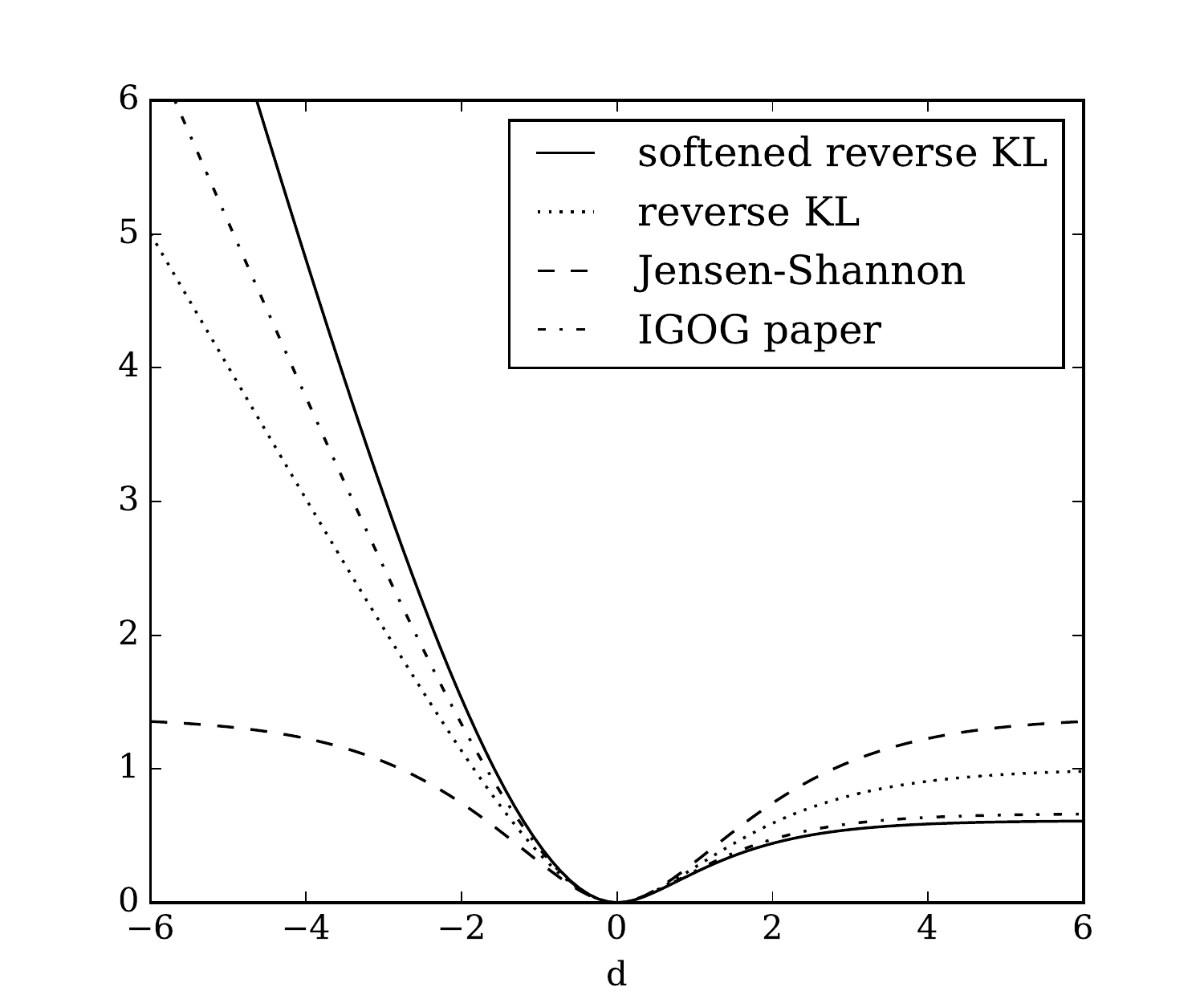}
  \caption{%
    Plots of $s_f(d)$ for various reverse KL-like f-divergences and Jensen-Shannon.
    Softened reverse KL is the divergence effectively minimized by non-saturating GAN training.
    IGOG is the divergence derived by \citet{poole2016improved}.
  }%
  \label{fig:symm_srkl}
\end{figure}
\figref{fig:symm_srkl} shows the symmetry-preserving representation $s_f(d)$ for
Jensen-Shannon and softened reverse KL, as well as the reverse KL for comparison.
The qualitative behavior of softened reverse KL is quite similar to reverse KL.
The softened version has a steeper slope in the roughly linear left tail and
changes the right tail behavior slightly, but these are relatively minor differences.
Jensen-Shannon is extremely different to reverse KL and softened reverse KL.

Tail weights and boundedness provide a very concise way to see the qualitative
behavior of non-saturating GAN training.
The softened reverse KL divergence has tail weights (2, 0), and so is unbounded, is
likely to have a strong gradient starting from a random initialization where large
mismatches are present, and penalizes left mismatches strongly but tolerates large
right mismatches and so is mode-seeking.
In contrast the Jensen-Shannon divergence approximately minimized by saturating GAN training
has tail weights (1, 1), and so is bounded, is likely to have a weak gradient in the
presence of large mismatches, and tolerates large left and right mismatches.

\subsection{Empirical pushforward plots}
\begin{figure}[t]
  \centering
  \includegraphics[trim=2mm 3mm 2mm 2mm,clip,scale=0.58]{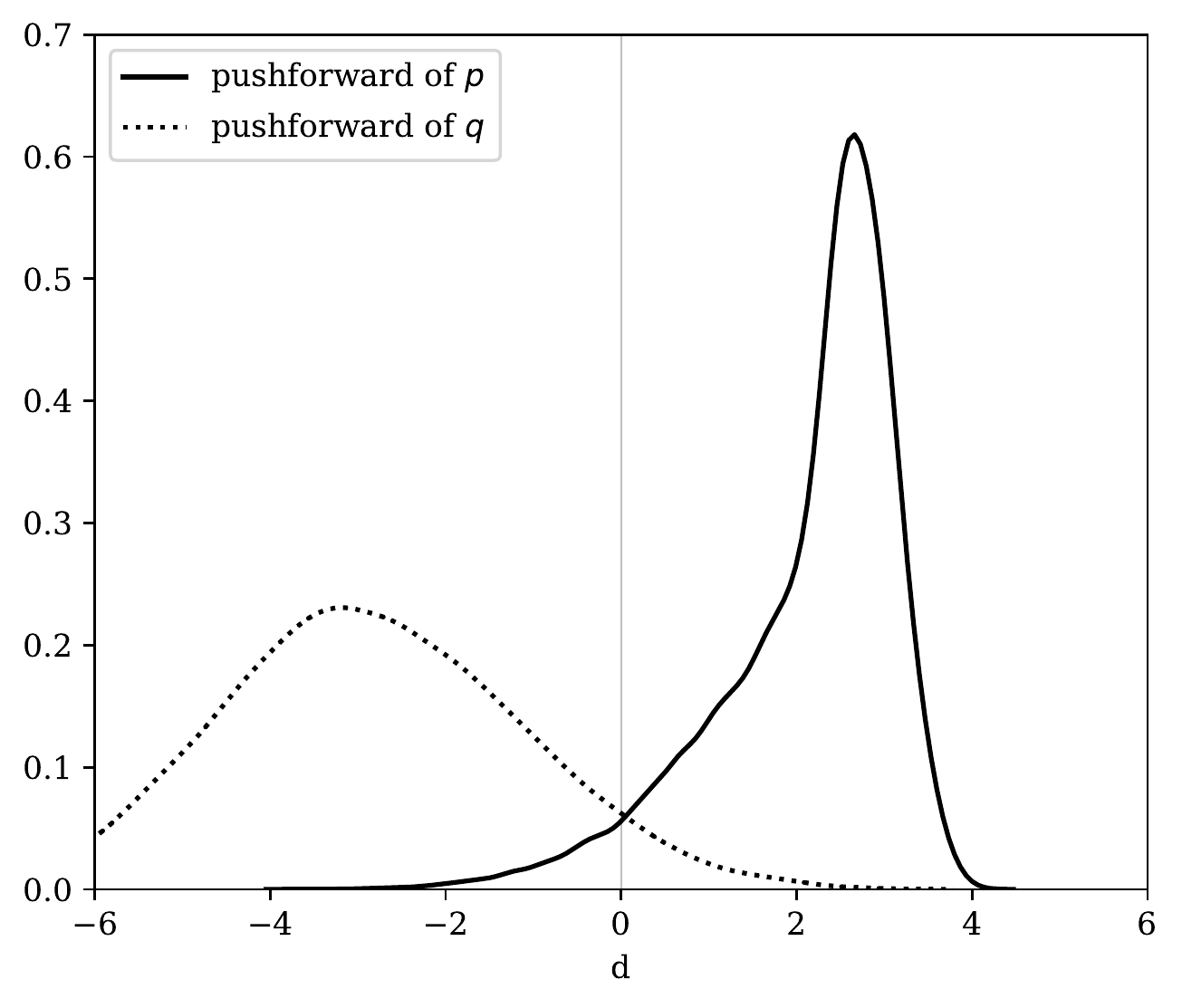}
  \caption{%
    Pushforward plot for StyleGAN \citep{karras2019style}.
    This shows the histogram of the critic output value on real and fake images.
    We can see that the critic is easily able to distinguish real and fake images
    most of the time, but we do not see substantial mode-seeking or covering behavior.
  }%
  \label{fig:pushforward-stylegan}
\end{figure}
In \sref{sec:pushforward-symm} we described pushforward plots obtained by pushing the
distributions $p$ and $q$ through the optimal critic $d^*$.
This is easy to approximate empirically, using the learned critic instead of the optimal
critic, by plotting the histogram of the critic output $d(x)$ for real and fake $x$.
Critic output histograms were previously investigated by \citet[Figure 1]{grewal2017variance}.
StyleGAN \citep{karras2019style} is a recent high-quality image generation GAN trained
with the non-saturating training scheme modified with a gradient penalty.
Its pushforward plot is shown in \figref{fig:pushforward-stylegan}.%
\footnote{%
  Pre-trained model \texttt{stylegan-ffhq-1024x1024.pkl} available at
  \url{https://github.com/NVlabs/stylegan}.
  The critic was trained for a further $6000$ examples while keeping the
  (non-moving-average) generator fixed.
}
From this plot we can read off the prevalence of different severities of left and right
mismatch, as well as compute the approximate value of any f-divergence.
Here we do not see substantial mode-seeking or covering:
there are no heavy left or right tails (as in \figref{fig:pushforward-axis-rkl} and
\figref{fig:pushforward-axis-kl}), and no separate peaks at large negative or
positive values of $d$ (as in \figref{fig:pushforward-cog-rkl} and
\figref{fig:pushforward-mnist-mode-collapse}).
More examples of pushforward plots are given in \sref{sec:more-pushforward-plots}.

\subsection{Previous discussion of non-saturating scheme}
\label{sec:previous-work}
The training dynamics of the non-saturating scheme and whether it can be motivated in a
principled way have been a source of discussion and some confusion.
In this section we review previous attempts to view non-saturating GAN training as a form of
divergence minimization.

\citet[Section 3]{goodfellow_generative_2014} claim that, compared to the saturating training
scheme based on the Jensen-Shannon divergence, the non-saturating training scheme
``results in the same fixed point of the dynamics of $G$ and $D$ but provides much stronger
gradients early in learning.''
It is true that the original and non-saturating generator gradients give the same
final result in the non-parametric case where $q$ is unrestricted, but this is fairly
trivial since both gradients lead to $q = p$, as do all divergences.
It is even true that the dynamics of training are essentially the same for the original
and non-saturating gradients when $q \approx p$, but again this is fairly trivial since all
f-divergences agree in this regime, as discussed in \sref{sec:f-gan-properties}.
However the ``fixed point of the dynamics'' is certainly not the same in the general case
of parametric $q$ with some model mismatch (see \sref{sec:experiments} for an empirical
demonstration).
Our results provide a precise way to view the relationship between saturating and
non-saturating generator gradients: they are optimizing different f-divergences.

\citet[Section 3.2]{nowozin_f-gan:_2016}%
\footnote{%
  In the NIPS paper but not in the arXiv preprint.
}
present a simple argument that the ``non-saturating'' f-GAN training
scheme has the same fixed points and that the original and non-saturating generator
gradients have the same direction.
However this argument is erroneous.
It is true that if $p \approx q$ then $(f^*)'(f'(u))$ is approximately $1$ everywhere,
and so the original and non-saturating generator gradients are approximately equal,
but this is true of any f-divergence.
There is no guarantee that the regime $p \approx q$ will ever be approached in the general
case where $q$ belongs to a parametric family, and it is not the case that the original and
non-saturating generator gradients point in approximately the same direction in general
(see \sref{sec:experiments} for an empirical demonstration).
In fact, the non-saturating form of generator gradient can have completely different
qualitative behavior from the original form.
For example, we showed in \sref{sec:equiv-ns} that using the non-saturating variant of the KL
generator gradient in fact optimizes reverse KL.

\citet{fedus_many_2018} argue against the view of GAN training as divergence minimization, based
in part on an empirical observation involving the non-saturating scheme.
They show \citep[Figure 2]{fedus_many_2018} that for two well-separated 1D Gaussians,
the non-saturating GAN loss has a large gradient even for the optimal critic, while Jensen-Shannon
has a very small gradient, and use this to argue against viewing non-saturating GAN training as
minimizing Jensen-Shannon.
The results presented in \sref{sec:f-gan-training} and \sref{sec:ns} show that this
set-up computes the gradient of the softened reverse KL, not Jensen-Shannon, and so this
observation is not on its own an argument against viewing non-saturating training as
divergence minimization.

\citet[Section 2.2.2]{arjovsky_towards_2017} show that non-saturating GAN training approximately
minimizes an objective function.
The objective function derived there is expressed as $\KLD{q}{p} - 2 \JS(p, q)$, which is a
rearrangement of the expression $2 \KLD{\half p + \half q}{p}$ derived in \sref{sec:equiv-ns}.
The paper suggests the negative sign of the second term is ``pushing for the
distributions to be different, which seems like a fault in the update'',
whereas writing the objective function as a divergence makes it clear this is not an issue.

\citet{poole2016improved} present a very similar view to that presented in this paper, including
recognizing that the generator and critic may be trained to optimize different f-divergences and
interpreting classic non-saturating GAN training as a scheme of this form.
The divergence derived there can be written as $D_f(p, q) = 2 \KLD{m}{p} + \KLD{p}{m}$ where
$m = \half p + \half q$ and has $f''(u) = u^{-2} - (1 + u)^{-2} = \frac{2 u + 1}{(1 + u)^2 u^2}$.
We refer to this as the \emph{improved generator objectives for GANs (IGOG)} divergence.
In contrast the softened reverse KL divergence $4 \KLD{m}{p}$ derived in \sref{sec:equiv-ns}
has $f''(u) = \frac{2}{u^2 (1 + u)}$.
We can see from \figref{fig:symm_srkl} and \tabref{tab:tail-weights} that the IGOG divergence is
qualitatively similar to the softened reverse KL derived in this paper, but they are not
identical.
The discrepancy between the two results is related to matching the value instead of the gradient.
In the notation developed in \sref{sec:f-gan-formulation}, \citet[Equation (8)]{poole2016improved}
consider the approximation
\begin{equation*}
  \expect_{x \sim q}\left[ f(p(x) / q(x)) \right]
    \approx \expect_{x \sim q}\left[ f(\exp(d(x))) \right]
\end{equation*}
and show that for the IGOG divergence the gradient of the right side is equal to the
non-saturating generator gradient.
This is a valid approximation of the value: for the optimal critic,
$d(x) = \log p(x) - \log q(x)$, so the left and right side have the same value.
However the gradients are not the same: the partial derivative of the left side
with respect to the parameters of $q$ involves two terms, one for each occurrence of
$q(x)$, and the partial derivative of the right side only includes one of these.
Thus non-saturating GAN training does not minimize the IGOG divergence.

\subsection{Practical implications}
The focus of this paper is theoretical, but a number of practical implications follow naturally:
\begin{itemize}
\item
  The mode-seeking behavior often observed during non-saturating GAN training
  is well explained by the softened reverse KL divergence having (2, 0) tail weights.
  If this is undesirable then using a heavier right tail may encourage diversity.
\item
  Our analysis suggests reverse KL and softened reverse KL are qualitatively similar.
  Given that reverse KL has a very slightly heavier right tail and better understood
  theoretical properties, it may make sense to use hybrid (reverse KL, JS) training
  (\sref{sec:f-gan-properties}) as a replacement for conventional non-saturating training.
\item
  To monitor progress during non-saturating GAN training, it makes sense to plot the softened
  reverse KL divergence rather than any other loss.
  For example, the conventional non-saturating generator loss
  $-\expect_z[ \log\sigma(d(g_\lam(z))) ]$
  only corresponds to the second term on the right side of \Eqref{eq:f-bound-def} and so would
  not be expected to decrease even if training is progressing well.
\item
  Hybrid training (\sref{sec:f-gan-properties}) shows that a single critic can be used to
  approximate any f-divergence.
  Plotting KL and reverse KL during training helps to monitor mode collapse and sample
  quality respectively.
  Pushforward plots are also useful, providing a fine-grained view of
  the prevalence and severity of mismatches between $p$ and $q$.
\item
  A critic trained with a symmetric divergence such as Jensen-Shannon will devote equal
  modelling effort to detecting left and right mismatches, and has the potential to detect
  mode collapse even if the generator is not able (due to optimization difficulties) or
  not strongly motivated (due to mode-seeking properties of the f-divergence used) to
  address it.
  This will be evident in both the pushforward plot and estimated KL.
  We provide an example of this in \sref{sec:more-pushforward-plots}.
\item
  The notion of tail weight may be useful for designing new GAN training algorithms.
  For example, the right tail weight could be annealed from $3/2$ to $1$ as training
  progresses.
  This would encourage diversity and help prevent mode collapse early in training while
  prioritizing sample quality later in training.
\end{itemize}
These suggestions for GAN training and avenues for future work are made possible
in part by the improved theoretical
understanding developed in this paper.

\clearpage
\bibliographystyle{apalike}
\bibliography{ns_gan}

\newpage
\onecolumn
\appendix
\section{Experimental validation}
\label{sec:experiments}
In order to validate our mathematical conclusions we performed GAN training on a simple
toy problem where we could explicitly plot the evolution of the generator parameters
during training.
In this simple set-up it is easy to verify that original, saturating GAN training scheme does
indeed converge to a minimum of the Jensen-Shannon divergence and that non-saturating GAN training
converges to a minimum of the softened reverse KL divergence.

The details of the experimental set-up were as follows.
The true distribution is $p(x) = 0.5\,\normal(x; 0, 0.3^2) + 0.5\,\normal(x; 2, 1)$, a mixture of two
1D Gaussians.
The generator is $q(x) = \normal(x; \mu, \sigma^2)$ and is parameterized by $\mu$ and $\sigma$,
both set initially to $1.8$.
The critic architecture in terms of Keras layers is: Dense(20), ELU, Dense(20), ELU, Dense(1).
We also tried training using the analytically optimal critic, which is equivalent to SGD on the
divergence value.
GAN training was performed using vanilla stochastic gradient descent.
In all cases the critic was trained using the canonicalized Jensen-Shannon divergence (learning
rate $2 \cdot 10^{-2}$).
The generator was trained with canonicalized Jensen-Shannon (learning rate $4 \cdot 10^{-3}$) as well as
the conventional non-saturating generator loss multiplied by two (to make it precisely equivalent
to softened reverse KL), reverse KL and the canonicalized IGOG divergence (all using a learning
rate of $2 \cdot 10^{-3}$).
A batch size of $256$ was used.
The critic was updated $5$ times between each generator step (alternating SGD).
Training proceeded for $4000$ generator updates.

\begin{figure*}
  \centering
  \includegraphics[trim=15mm 5mm 20mm 10mm,clip,scale=0.7]{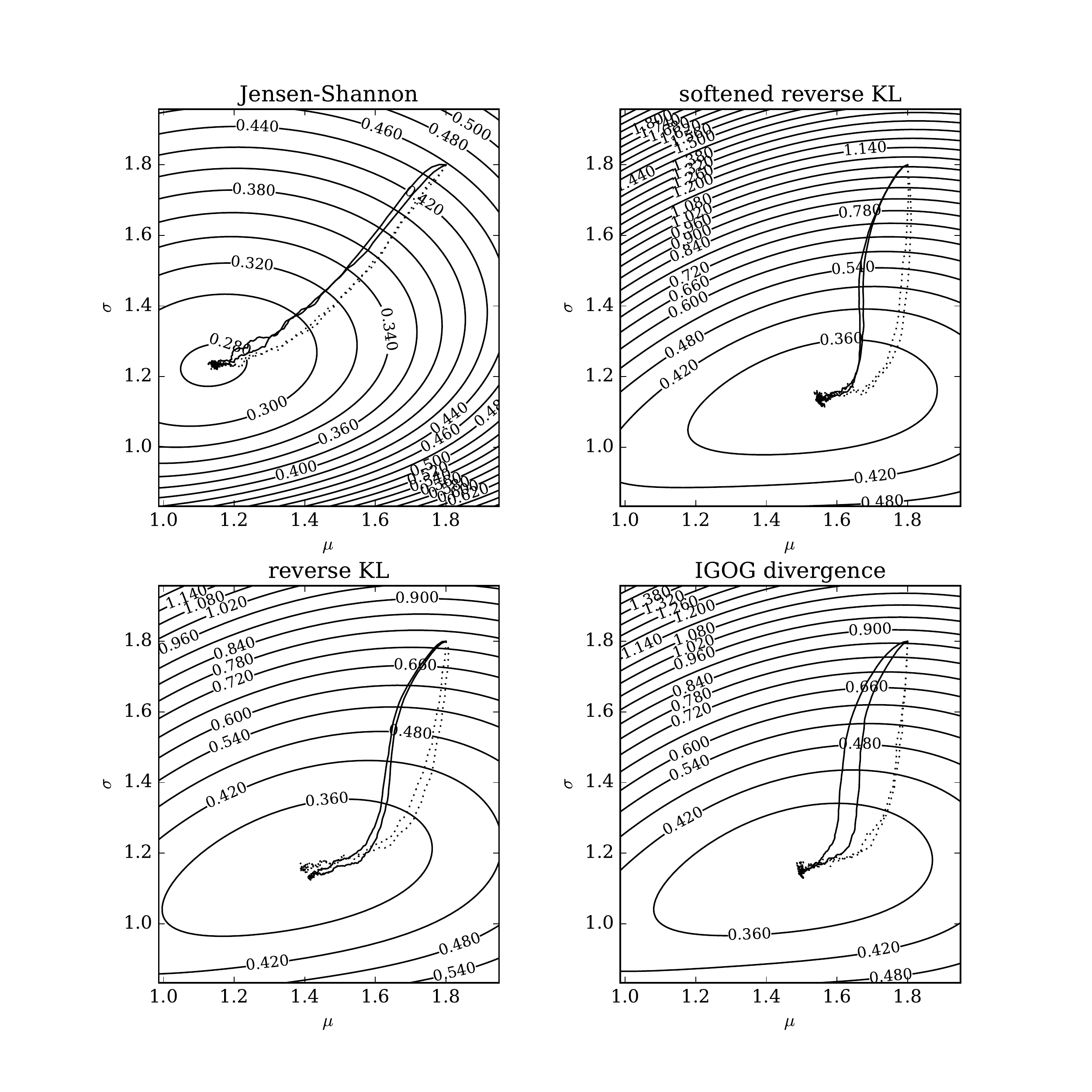}
  \caption{%
    Comparing training using the saturating and non-saturating GAN generator gradient on a toy
    problem.
    The true distribution $p$ is a mixture of two 1D Gaussians and the model distribution
    $q$ is a single Gaussian.
    Contour plots show the values of various divergences considered in this paper as a function of
    the generator parameters $\mu$ and $\sigma$.
    Lines show the progression of SGD-based GAN training using the original, saturating GAN
    training scheme (Jensen-Shannon panel), non-saturating GAN training (softened reverse KL panel),
    and hybrid (reverse KL, Jensen-Shannon) and (IGOG divergence, Jensen-Shannon) training schemes
    (bottom two panels).
    The solid lines show two independent training runs with a learned critic.
    The dotted lines show two independent training runs with the analytically optimal critic.
    Training approximately converges to the expected divergence minimum in all cases.
  }%
  \label{fig:contour-js-srkl}
\end{figure*}
The evolution of the generator parameters during training is shown in \figref{fig:contour-js-srkl}.
In all cases, training approximately converges to a minimum of the corresponding divergence value
as expected based on the theoretical arguments presented in the main body.

\section{Divergence boundedness}
Some f-divergences such as Jensen-Shannon are bounded, while others such as KL are
unbounded.
This is relevant for training since we might expect bounded divergences to have
weak gradients when starting from a random initialization.
In this section we provide a characterization of the boundedness of $D_f$ in terms of the
behavior of $f(u)$ for small and large $u$, following \citet{vajda1972f}.
This will be related to tail weight in the next section.

Given an f-divergence $D_f$, define:
\begin{align}
  M &= \sup_{p, q} D_f(p, q)
  \\
  \label{eq:left-bound-def}
  M_0 &= \sup_{u \in (0, 1)} f(u)
  \\
  M_\infty &= \sup_{u > 1} \frac{f(u)}{u} = \sup_{u \in (0, 1)} f_\text{R}(u)
\end{align}
The supremums are guaranteed to exist but may be infinite.
\citet[Theorem 2,``range of values'']{vajda1972f} showed that
\begin{equation}
  \label{eq:range-of-values}
  M = M_0 + M_\infty
\end{equation}
We refer to a divergence as \emph{left-bounded} if $M_0 < \infty$,
\emph{right-bounded} if $M_\infty < \infty$,
and \emph{bounded} if $M < \infty$.
Thus a divergence is bounded iff it is left-bounded and right-bounded.

We briefly describe how to establish \Eqref{eq:range-of-values}.
It is straightforward to see that $M \leq M_0 + M_\infty$:
on the right side of \Eqref{eq:f-left-right-decomp}, the first term has $f(u) \leq M_0$
and the second term has $f_\text{R}(u^{-1}) \leq M_\infty$, so
$D_f(p, q) \leq M_0 + M_\infty$ as desired.
For the converse direction, we need to exhibit probability distributions $p$ and $q$
with $D_f(p, q)$ arbitrarily close to $M_0 + M_\infty$.
This is most straightforward in the case where $p$ and $q$ are distributions over the
two-point set $\{0, 1\}$.
Given $r, s \in (0, 1)$, let $p_r(0) = r$ and $q_s(0) = s$.
This determines $p_r$ and $q_s$ entirely, and we have
\begin{equation}
    D_f(p_r, q_s)
      = s f\left(\frac{r}{s}\right) + (1 - r) f_\text{R}\left(\frac{1 - s}{1 - r}\right)
\end{equation}
Now a defining function $f$ is monotonically decreasing on $(0, 1)$, so we can replace
the supremum in \Eqref{eq:left-bound-def} with a limit: $f(u) \to M_0$ as $u \to 0$.
Thus as $r \to 0$ and $s \to 1$, $D_f(p_r, q_s) \to M_0 + M_\infty$ as desired.
For the continuous case where $p$ and $q$ are densities over $\reals^K$, we reduce this
to the discrete case by considering mixtures of Gaussians with shrinking covariances.
Fix any two distinct points $\mu_0, \mu_1 \in \reals^K$.
Let
$\tilde{p}_{r \sigma}(x)
    = r\,\normal(x; \mu_0, \sigma^2 I) + (1 - r)\,\normal(x; \mu_1, \sigma^2 I)$
and
$\tilde{q}_{s \sigma}(x)
    = s\,\normal(x; \mu_0, \sigma^2 I) + (1 - s)\,\normal(x; \mu_1, \sigma^2 I)$
where $r, s \in (0, 1)$ as before.
As $\sigma \to 0$, $D_f(\tilde{p}_{r \sigma}, \tilde{q}_{s \sigma}) \to D_f(p_r, q_s)$.
Given $\varepsilon > 0$, we can find $r, s \in (0, 1)$ such that $D_f(p_r, q_s)$ is within
$\varepsilon$ of $M_0 + M_\infty$, and given this $r, s$ we can find $\sigma > 0$ such
that $D_f(\tilde{p}_{r \sigma}, \tilde{q}_{s \sigma})$ is within $\varepsilon$ of
$D_f(p_r, q_s)$.
Thus $D_f(\tilde{p}_{r \sigma}, \tilde{q}_{s \sigma})$ is within $2 \varepsilon$ of
$M_0 + M_\infty$.
Thus there exist continuous probability distributions with divergence value arbitrarily
close to $M_0 + M_\infty$.

\section{Tail weight properties and computation}
\label{sec:tail-weights-more}
In this section we provide more details on tail weights.
We describe the relationship between boundedness and tail weights and provide some examples
of how tail weights may be computed.

Tail weight determines boundedness.
It can be checked by integrating and bounding that a divergence with $(L, R)$ tail
weights is left-bounded iff $L < 2$ and right-bounded iff $R < 2$.
Thus $D_f$ is bounded iff $L, R < 2$.
Boundedness properties can also be seen in \figref{fig:symm}.
Left and right boundedness of $D_f$ is equivalent to left and right boundedness of $s_f$.
Thus we can see that reverse KL is left-unbounded but right-bounded, for example.

Tail weights also interact in a simple and intuitive way with linearity:
if one f-divergence has $(L_1, R_1)$ tail weights and another has $(L_2, R_2)$ tail weights
then their sum has $(\max_i L_i, \max_i R_i)$ tail weights.

Tail weights are relatively straightforward to compute for a given f-divergence.
The definition from \sref{sec:tail-weights} is that $D_f$ has left tail weight $L$ if
$f''(u) \sim C u^{-L}$ as $u \to 0$ for some $C \in \reals$.
Here $\sim$ may be read as ``is asymptotic to'' and ``$g(u) \sim h(u)$ as $u \to a$''
has the meaning ``$g(u) / h(u) \to 1$ as $u \to a$''.
Thus $D_f$ has left tail weight $L$ iff $u^L f''(u)$ tends to a constant as $u$ tends to
zero.
The general idea for computing left tail weight is therefore to work out what power of $u$
we need to multiply $f''(u)$ by in order for it to tend to a constant as $u$ tends to zero.
The $\sim$ notation makes this a little easier since we can make use of two properties:
if $f(u) \sim g(u)$ and $g(u) \sim h(u)$ as $u \to a$ then $f(u) \sim h(u)$; and
if $f_1(u) \sim g_1(u)$ and $f_2(u) \sim g_2(u)$ as $u \to a$ then
$f_1(u) f_2(u) \sim g_1(u) g_2(u)$.
For example, Jensen-Shannon has $f''(u) = \frac{2}{u (1 + u)}$.
As $u \to 0$, we have $\frac{2}{1 + u} \to 2$, and so $\frac{2}{1 + u} \sim 2$, and so
$f''(u) \sim 2 u^{-1}$, and so the left tail weight is $1$.
For the softened reverse KL, $f''(u) = \frac{2}{u^2 (1 + u)}$, so in the same way we see that
$f''(u) \sim 2 u^{-2}$ and so the left tail weight is $2$.
The right tail weight is defined as $R$ if $f''(u) \sim D u^{R - 3}$ as $u \to \infty$
for some $D \in \reals$.
For example, for Jensen-Shannon we can multiply the top and bottom by $u$ to obtain
$f''(u) = \frac{2 u}{u^2 (1 + u)}$, and $\frac{u}{1 + u} \to 1$ as $u \to \infty$, so
$f''(u) \sim 2 u^{-2}$, so $R - 3 = -2$, so $R = 1$.
Alternatively we can replace $u$ by $u^{-1}$ and seek the asymptotic behavior as
$u \to 0$.

\section{Examples of pushforward plots}
\label{sec:more-pushforward-plots}
\begin{figure*}
  \centering
  \includegraphics[trim=10mm 0mm 10mm 10mm,clip,scale=0.6]{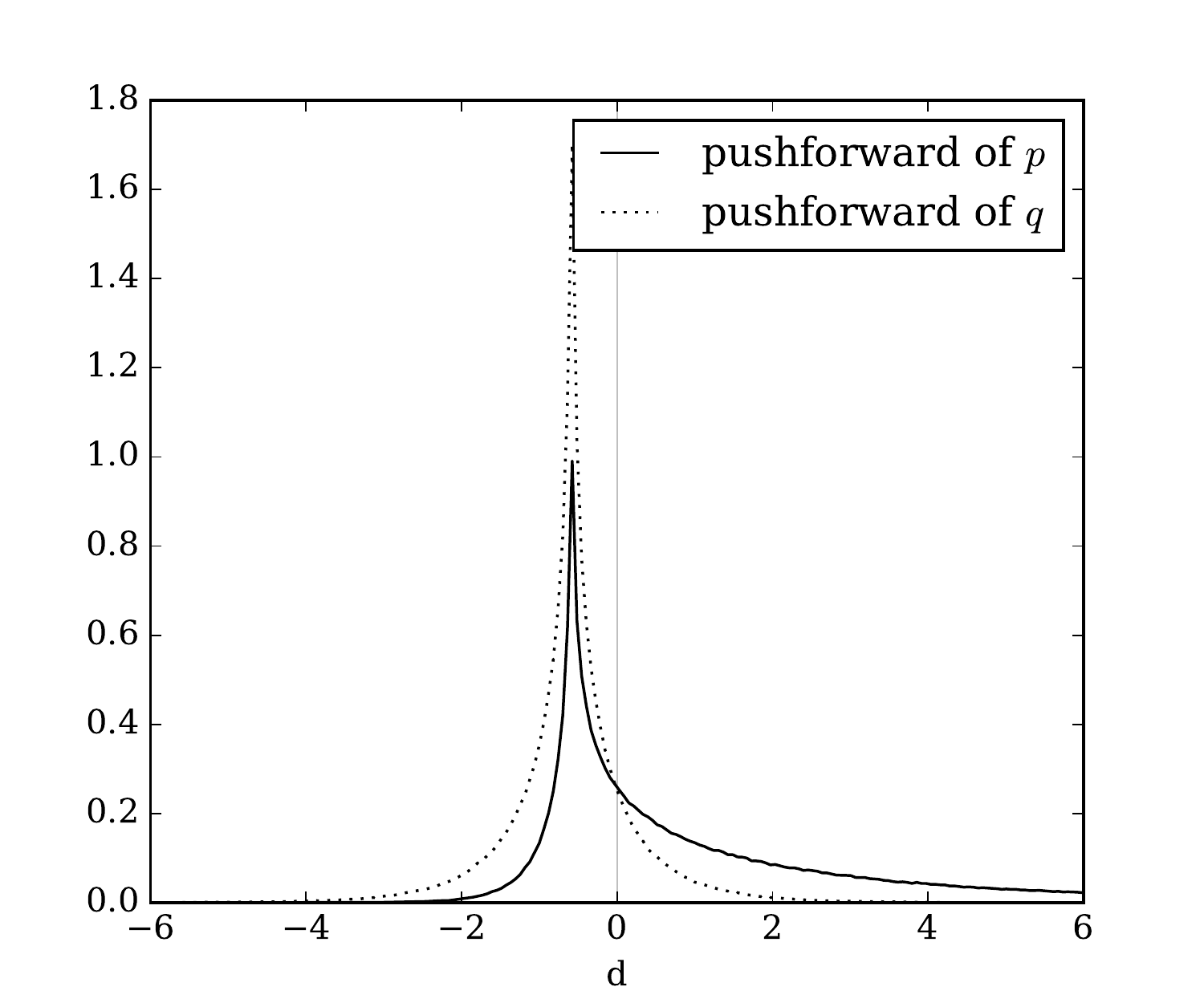}
  \includegraphics[trim=10mm 0mm 15mm 10mm,clip,scale=0.6]{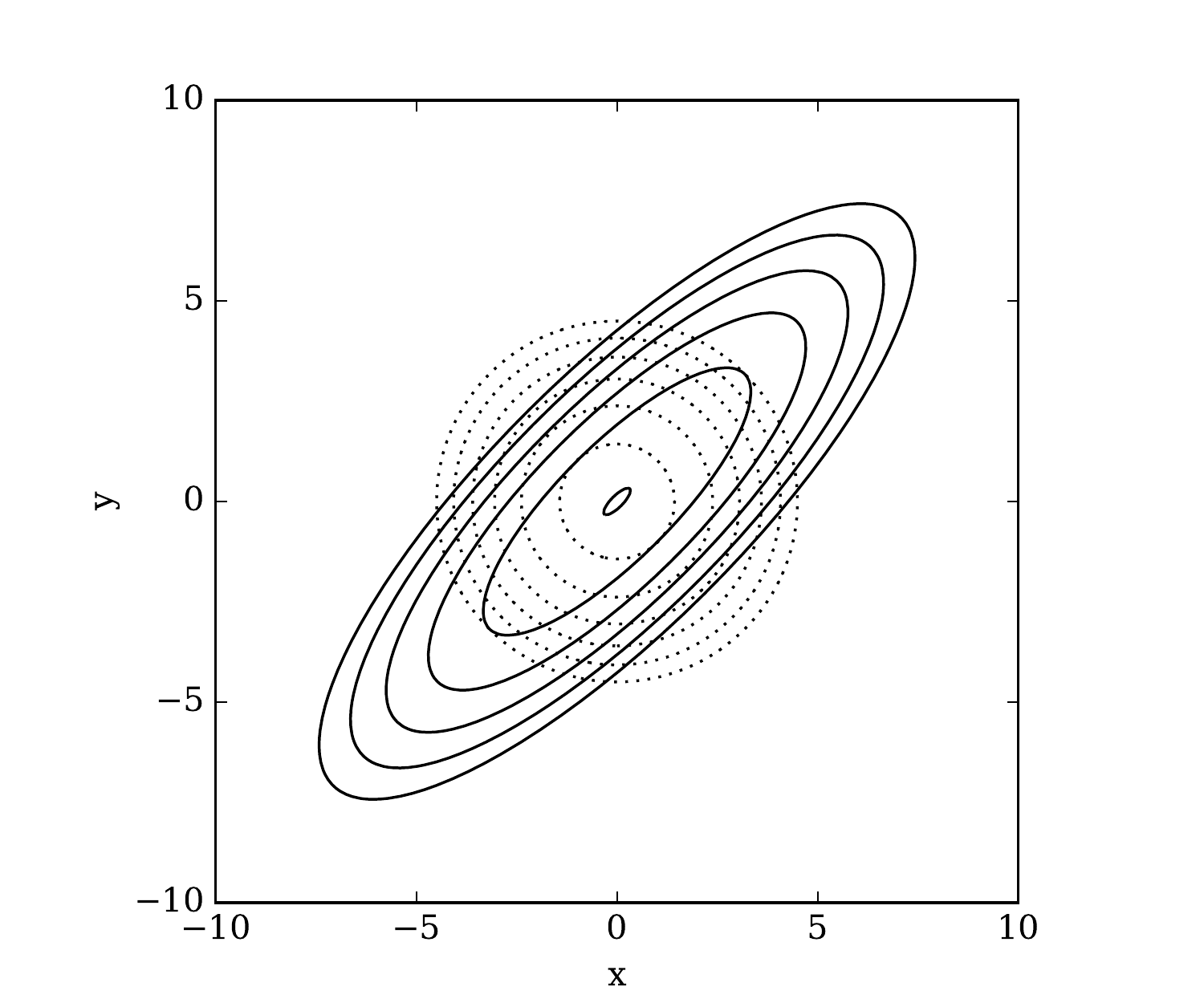}
  \caption{%
    Plots of the pushforward densities $\tilde{p}(d)$ and $\tilde{q}(d)$ for the case where
    $p$ is a 2D Gaussian with covariance $[[5.5, 4.5], [4.5, 5.5]]$ and $q$ is a diagonal covariance
    model fit with reverse KL.
    We can see from the heaviness of the tails that there are substantial right mismatches
    and that $q$ is more compact than $p$.
    A contour plot of the data-space density of $p$ and $q$ is also shown.
  }%
  \label{fig:pushforward-axis-rkl}
\end{figure*}
\begin{figure*}
  \centering
  \includegraphics[trim=10mm 0mm 10mm 10mm,clip,scale=0.6]{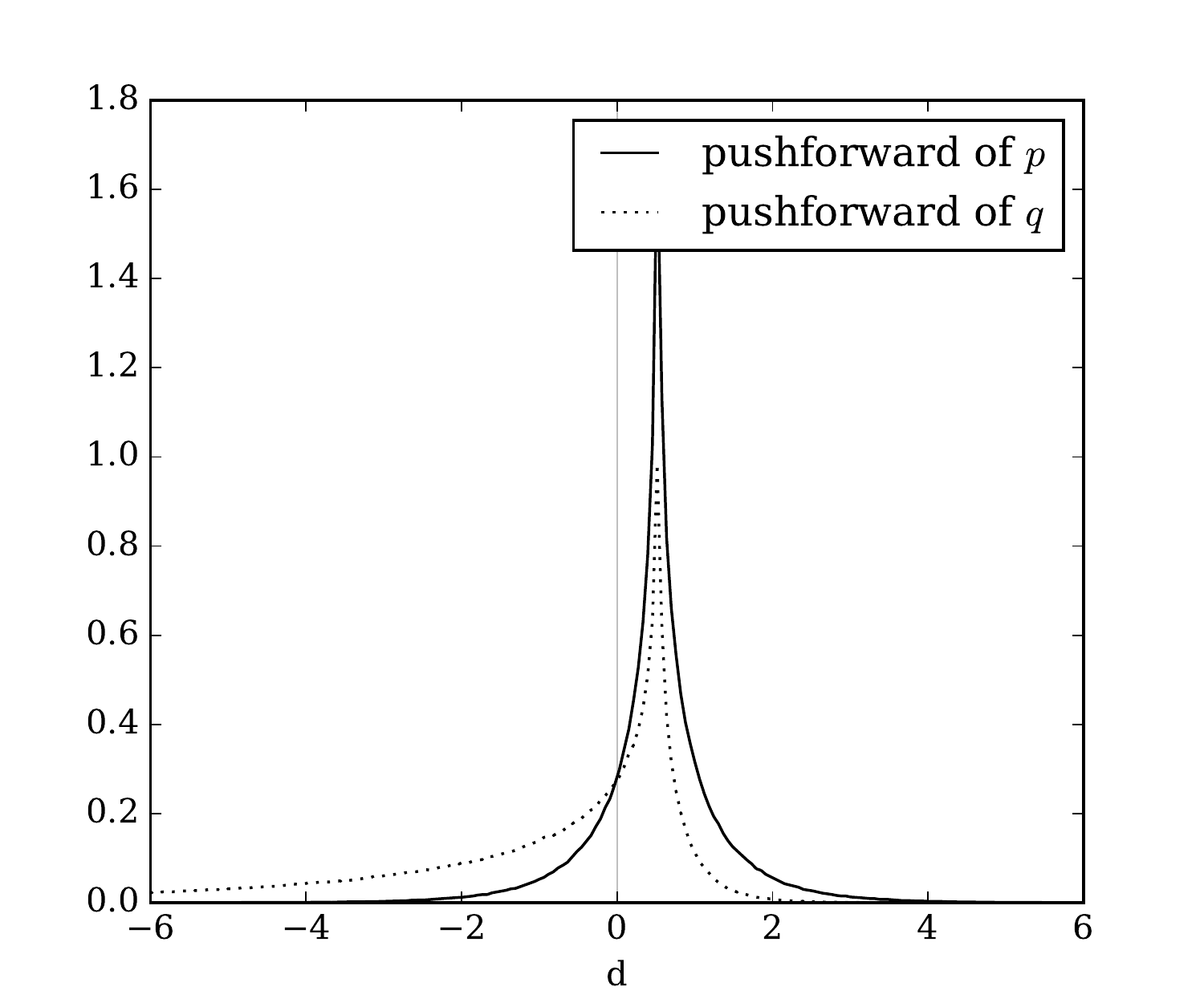}
  \includegraphics[trim=10mm 0mm 15mm 10mm,clip,scale=0.6]{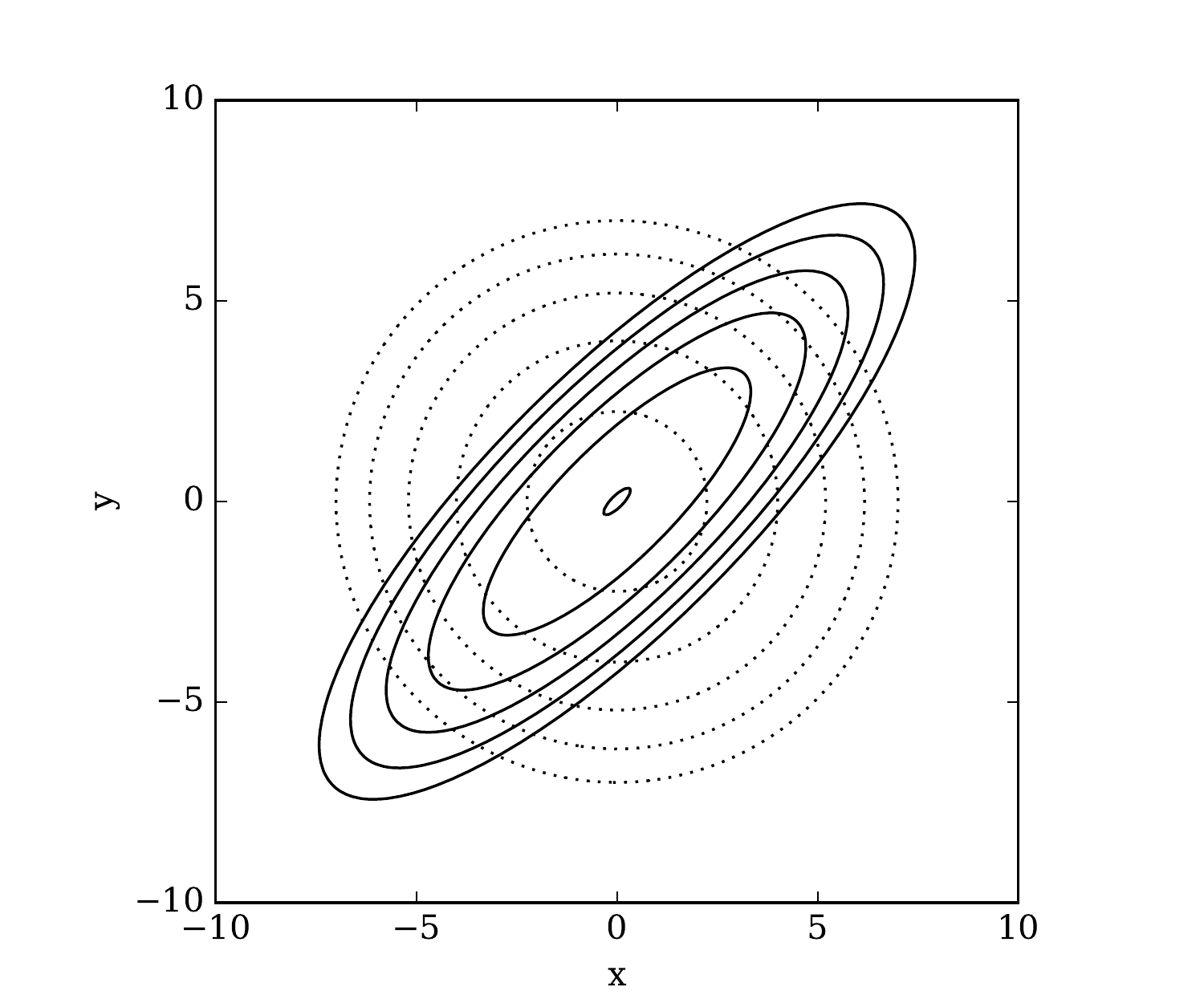}
  \caption{%
    The analogue of \figref{fig:pushforward-axis-kl} for KL instead of reverse KL.
    We can see from the heaviness of the tails that there are substantial left mismatches
    and that $q$ is broader than $p$ (covering behavior).
  }%
  \label{fig:pushforward-axis-kl}
\end{figure*}
\begin{figure*}
  \centering
  \includegraphics[trim=10mm 0mm 10mm 10mm,clip,scale=0.6]{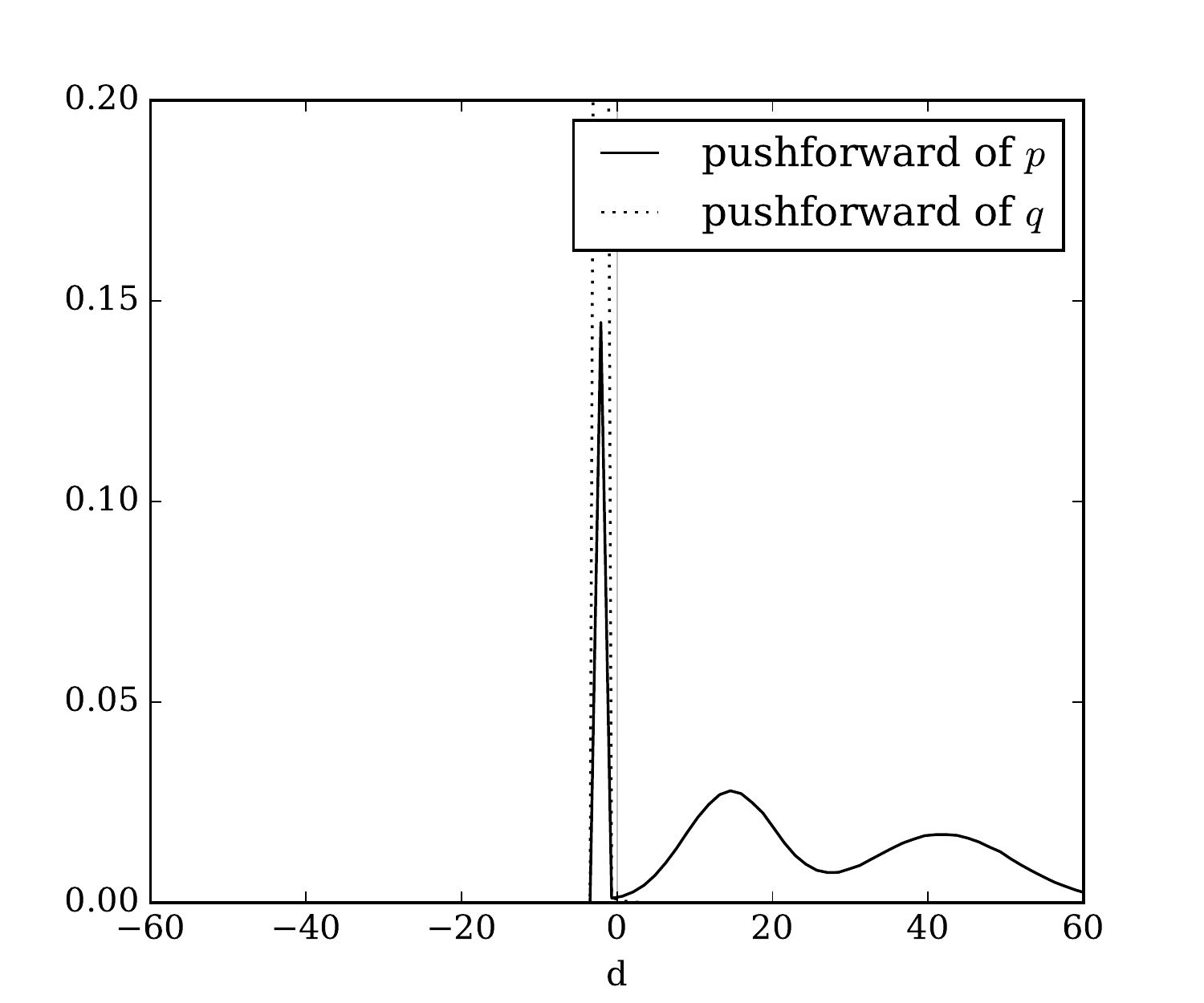}
  \includegraphics[trim=10mm 0mm 15mm 10mm,clip,scale=0.6]{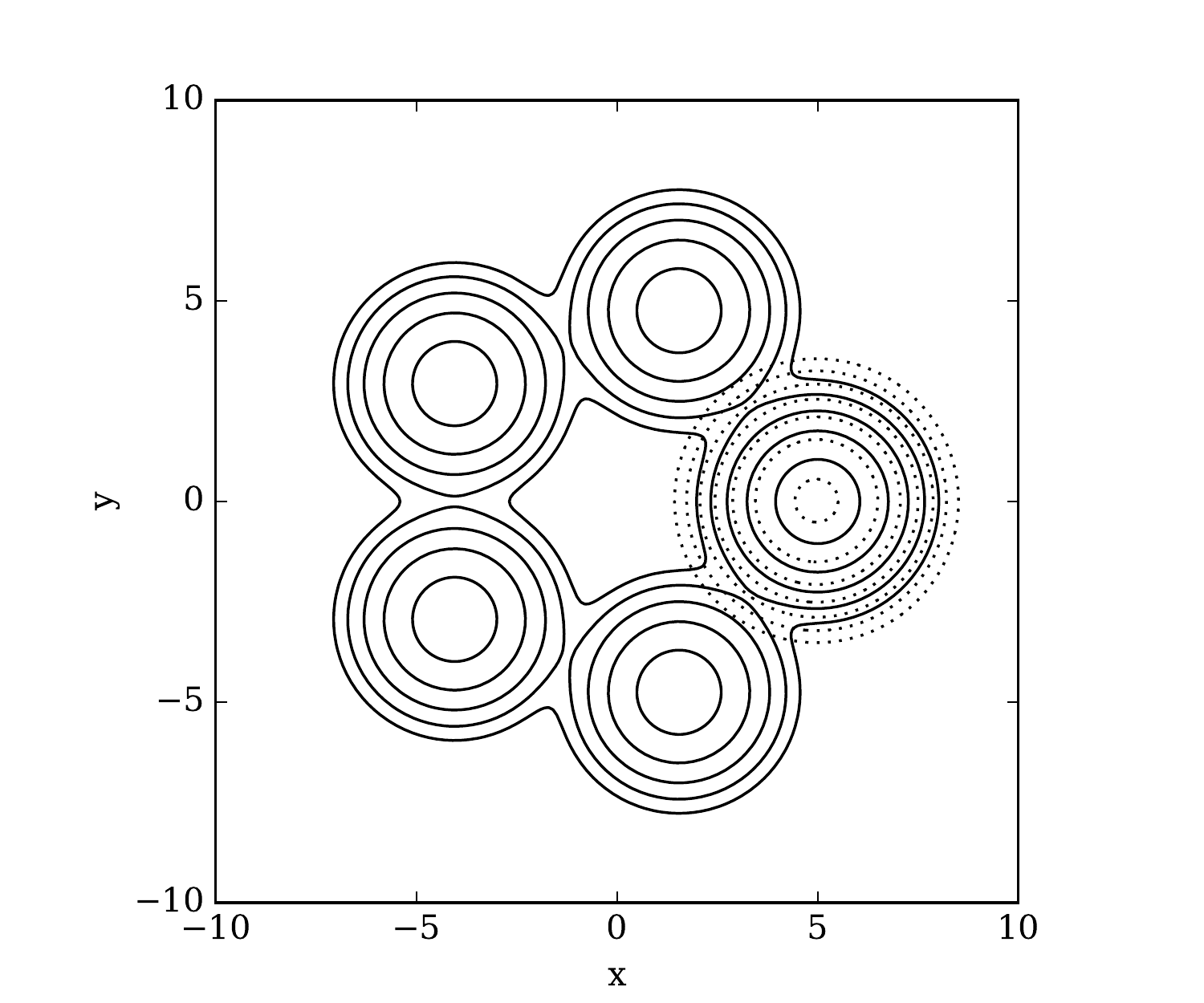}
  \caption{%
    Plots of the pushforward densities $\tilde{p}(d)$ and $\tilde{q}(d)$ for the case where
    $p$ is a 2D mixture of five Gaussians with means arranged in a circle at radius $5$ from
    the origin and identity covariance matrices and $q$ is an isotropic Gaussian with mean
    and scale fit using reverse KL.
    This is a prototypical example of mode collapse.
    We can actually see the modes which are being dropped in the pushforward plot:
    there is one peak in $\tilde{p}(d)$ for the two nearer dropped modes and one peak for
    the two further dropped modes.
    Note the wider scale of $d$.
    A contour plot of the data-space density of $p$ and $q$ is also shown.
  }%
  \label{fig:pushforward-cog-rkl}
\end{figure*}
\begin{figure*}
  \centering
  \includegraphics[trim=10mm 0mm 10mm 10mm,clip,scale=0.6]{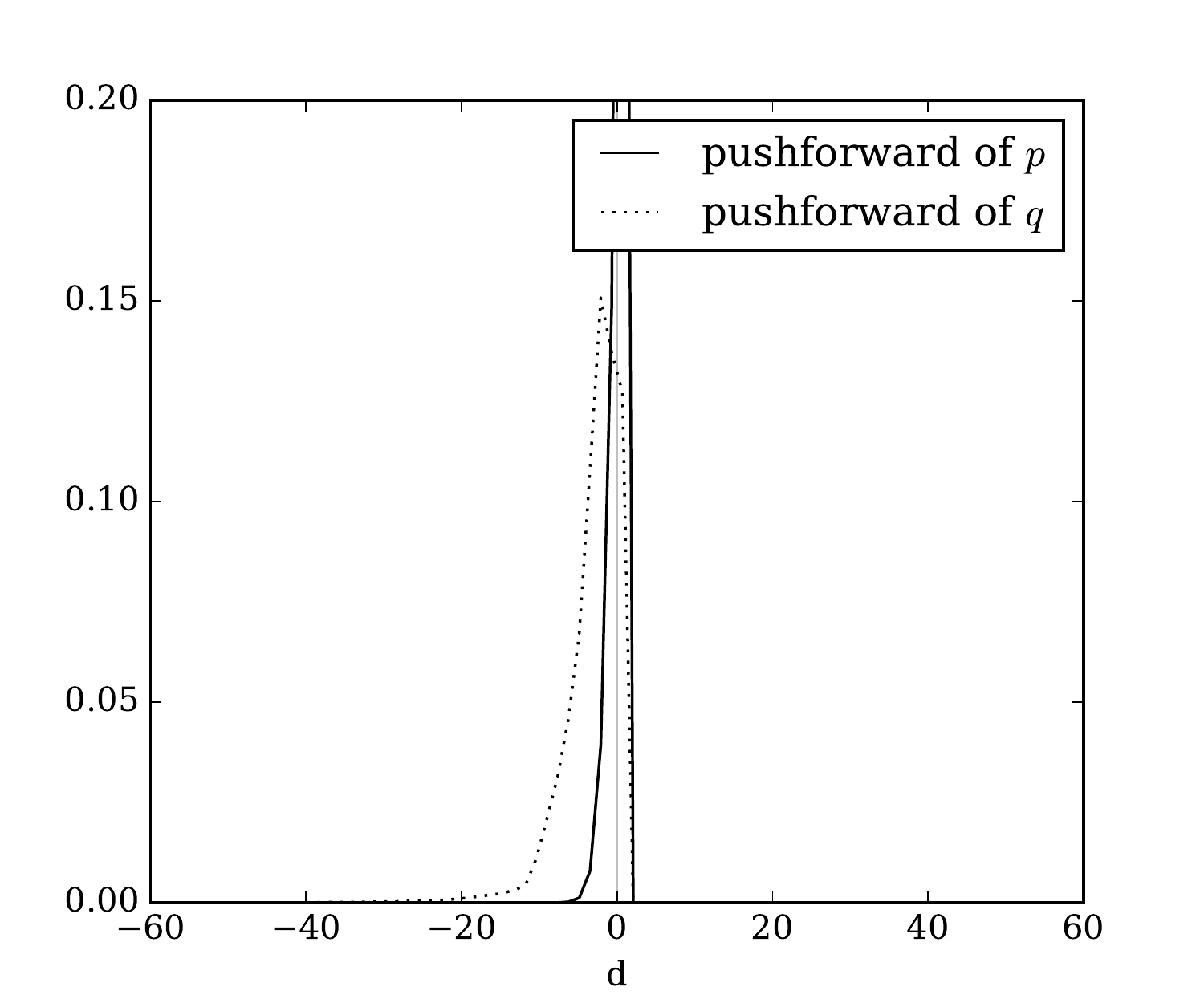}
  \includegraphics[trim=10mm 0mm 15mm 10mm,clip,scale=0.6]{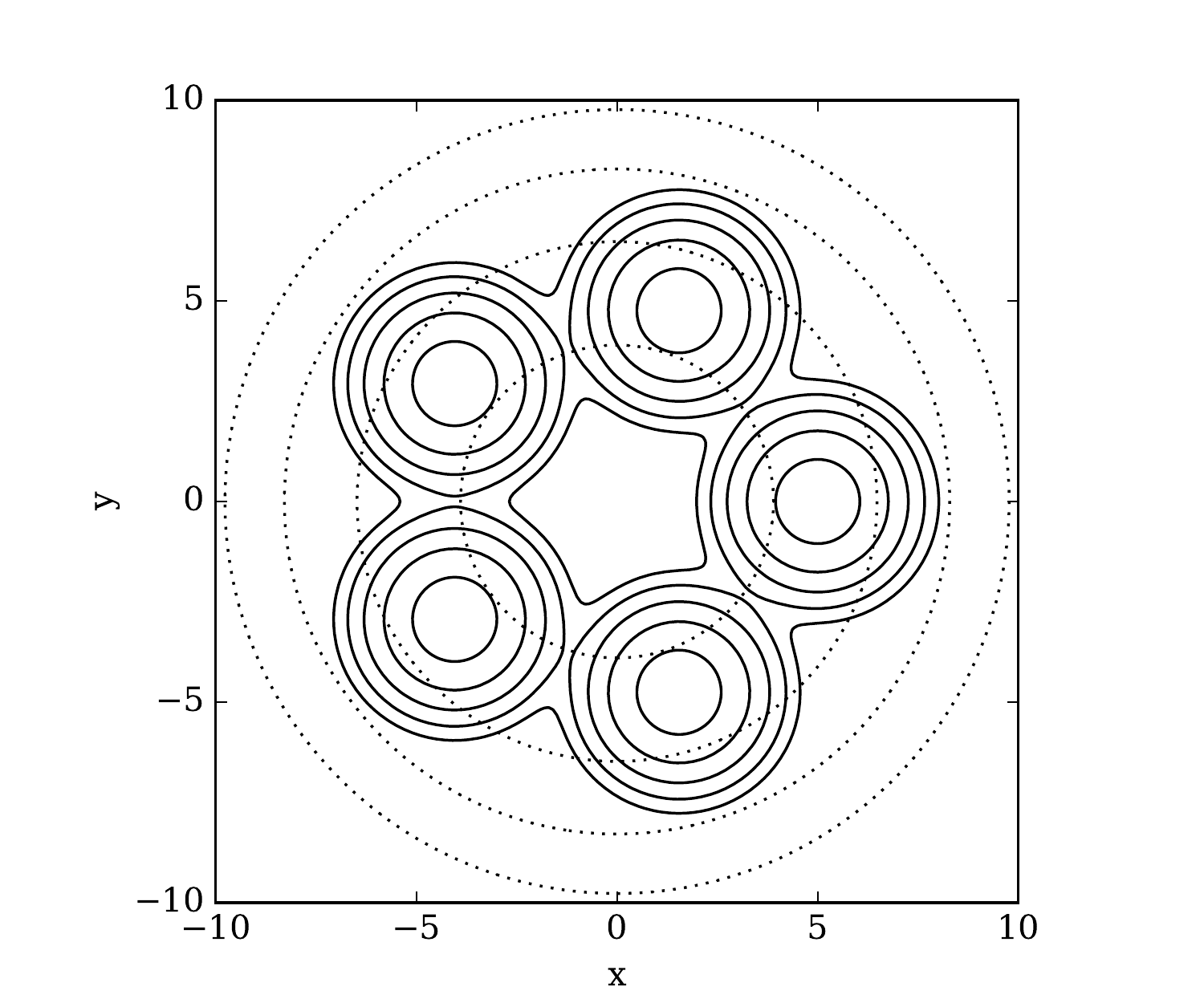}
  \caption{%
    The analogue of \figref{fig:pushforward-cog-kl} for KL instead of reverse KL.
    We can see the covering behavior in the pushforward plot by the relatively strong
    left tail of $\tilde{q}(d)$, though in this case the left mismatches caused by
    fitting with KL are less extreme than the right mismatches caused by fitting with
    reverse KL.
  }%
  \label{fig:pushforward-cog-kl}
\end{figure*}
\begin{figure*}
  \centering
  \includegraphics[trim=10mm 0mm 15mm 10mm,clip,scale=0.6]{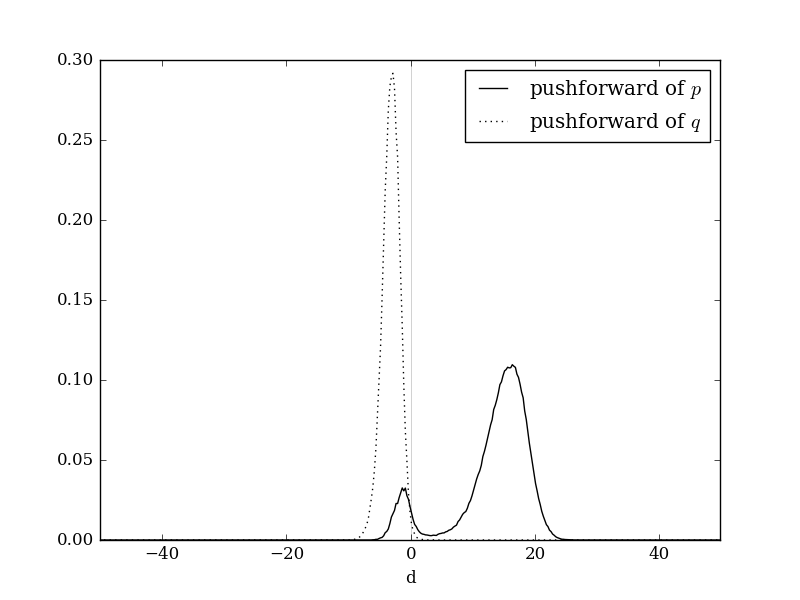}
  \includegraphics[trim=0mm 0mm 0mm 0mm,clip,scale=0.6]{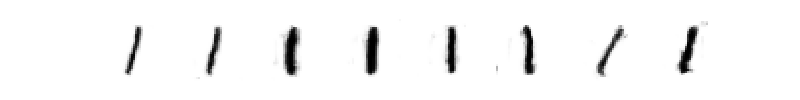}
  \caption{%
    Example of a pushforward plot demonstrating mode collapse on a real dataset.
    Samples from the generator are shown below the pushforward plot.
    The generator has a StyleGAN-like architecture and was trained on MNIST digits using a hybrid
    (reverse KL, Jensen-Shannon) training scheme.
    The generator output lacks diversity, consisting of various forms of the digit $1$.
    This is evidenced by a large right mismatch in the pushforward plot representing the nine other
    digits which have extremely low probability of being produced by the generator.
    Here the learned critic rather than the optimal critic is used for the pushforward.
    It is interesting to note that the critic has learned to detect the mode collapse (that is,
    learned to assign a score indicating a large mismatch to natural images of the other nine digits),
    even though the generator training has not been able to use this information effectively,
    presumably due to optimization difficulties and the use of a divergence which does not penalize
    right mismatches heavily.
  }%
  \label{fig:pushforward-mnist-mode-collapse}
\end{figure*}
In this section we show some examples of pushforward plots for a variety of learned generators.
This is intended to provide a more intuitive understanding of what pushforward plots capture,
as well as the ways in which they might be useful for monitoring training.

\figref{fig:pushforward-axis-rkl} and \figref{fig:pushforward-axis-kl} show mode-seeking and
covering behavior for a model trained with reverse KL and KL respectively.
Here the model mismatch comes from the generator using a diagonal covariance while the data
is generated from a full covariance Gaussian with strong correlation between the two dimensions.
\figref{fig:pushforward-cog-rkl} and \figref{fig:pushforward-cog-kl} show similar behavior for
a ``circle of Gaussians'' example.
Here the mismatch comes from the true distribution being a mixture while the generator is
unimodal.
\figref{fig:pushforward-mnist-mode-collapse} shows an example of mode collapse for a generator
trained on MNIST digits.
It is interesting to note that the critic is able to accurately detect the mode collapse, even
though the generator training has not been able / motivated to use this information to
remedy the problem.

\section{Divergence symmetrization and softening}
\label{sec:softening}
In this section we describe three simple operations that can be applied to an divergence
to obtain another divergence: reversing, symmetrizing and \emph{softening}.
Many common f-divergences can be obtained from others in this way.
For example, all the f-divergences considered in this paper can be obtained by applying
these operations to the KL divergence.
We refer to $4 \KLD{\half p + \half q}{p}$ as the \emph{softened reverse KL divergence}
because it may be obtained by applying the softening operation to the reverse KL.

Consider applying an operation to a divergence $D$ to obtain another divergence $\tilde{D}$.
If this operation maps f-divergences to f-divergences then we may also think of it as mapping
a defining function $f$ to another defining function $\tilde{f}$.
We already saw the reversing operation $D \mapsto \tilde{D}$ where $\tilde{D}(p, q) = D(q, p)$
in \sref{sec:mismatches}.
This has $\tilde{f} = f_\text{R}$ where $f_\text{R}(u) = u f(u^{-1})$.
In this case
\begin{equation}
  \tilde{f}''(u) = f_\text{R}''(u) = u^{-3} f''(u^{-1})
\end{equation}
and $a_{f_\text{R}}(d) = -b_f(-d)$ and $b_{f_\text{R}}(d) = -a_f(-d)$ as might be
expected intuitively.
Symmetrization means $\tilde{D}(p, q) = \half D(p, q) + \half D(q, p)$.
This has $\tilde{f} = \half f + \half f_\text{R}$ and
\begin{equation}
  \tilde{f}''(u) = \half f''(u) + \half f_\text{R}''(u)
\end{equation}
We define \emph{(q-)softening} as replacing $q$ with $m = \half p + \half q$,
i.e.\ $\tilde{D}(p, q) = 4 D(p, m)$.
This has $\tilde{f}(u) = 2 (1 + u) f(\frac{2 u}{1 + u})$ and
\begin{equation}
  \tilde{f}''(u) = \frac{8}{(1 + u)^3} f''\left(\frac{2 u}{1 + u}\right)
\end{equation}
The factor of $4$ in the definition of $\tilde{D}$ ensures that a softened canonical
divergence remains canonical.
Softening has the potential to make large right mismatches much less severely penalized,
since in regions of space where $p(x) / q(x)$ was large because $p(x)$ was moderate
and $q(x)$ was tiny, $p(x) / m(x)$ is now approximately $2$, so a large right mismatch
is only penalized by the softened divergence as much as a moderate right mismatch is
penalized by the original divergence.
This is reflected in the tail weights: It is easy to show using the tools we have developed
above that if the original divergence has $(L, R)$ tail weights then the softened divergence
has $(L, 0)$ tail weights.
For completeness, we could also define \emph{p-softening} as replacing $p$ with
$m = \half p + \half q$, i.e.\ $\tilde{D}(p, q) = 4 D(m, q)$.
This has $\tilde{f}(u) = 4 f(\half(1 + u))$ and
\begin{equation}
  \tilde{f}''(u) = f''\left(\half (1 + u)\right)
\end{equation}
If $D_f$ has $(L, R)$ tail weights then the p-softened divergence has $(0, R)$ tail
weights.
Softening was considered by \citet[Equations (41) and (42)]{vajda2009metric}, where it was
referred to as \emph{normalization}.

Many f-divergences can be written concisely as a series of these operations.
For example reverse KL is Reverse(KL), Jeffreys is Symmetrize(KL), the
canonicalized K-divergence $4 \KLD{p}{m}$ \citep{cha2007comprehensive} is Soften(KL)
and canonicalized Jensen-Shannon is Symmetrize(Soften(KL)).
The softened reverse KL divergence $4 \KLD{m}{p}$ that is the focus of this paper is
Soften(Reverse(KL)).

\end{document}